\documentclass[journal]{IEEEtran}
\usepackage{amsmath,amsfonts}
\usepackage{algorithm}
\usepackage{array}
\usepackage[caption=false,font=normalsize,labelfont=sf,textfont=sf]{subfig}
\usepackage{textcomp}
\usepackage{stfloats}
\usepackage{url}
\usepackage{verbatim}
\usepackage{graphicx}
\usepackage{cite}
\usepackage{accsupp}
\usepackage[accsupp]{axessibility} 
\usepackage{multirow}
\usepackage{makecell}
\usepackage{booktabs} 
\usepackage{enumitem}
\usepackage{algpseudocode}
\usepackage{float}
\usepackage[T1]{fontenc}
\usepackage[labelfont=bf]{caption}
\usepackage{ragged2e}
\usepackage{setspace}
\usepackage{hyperref}
\begin{document}

\title{Exploring What Why and How: \\
A Multifaceted Benchmark for Causation Understanding of Video Anomaly
}

\author{Hang Du, Guoshun Nan,~\IEEEmembership{Member, IEEE,} Jiawen Qian, Wangchenhui Wu, Wendi Deng, Hanqing Mu, Zhenyan Chen, Pengxuan Mao, Xiaofeng Tao,~\IEEEmembership{Senior Member, IEEE}, Jun Liu,~\IEEEmembership{Senior Member, IEEE}
\IEEEcompsocitemizethanks{
\IEEEcompsocthanksitem \rule{0.93\linewidth}{0.5pt}\protect\\
\textbullet \quad H. Du, G. Nan, J. Qian, W. Wu, W. Deng, H. Mu, Z. Chen and X. Tao are with National Engineering Research Center for Mobile Network Technologies, Beijing University of Posts and Telecommunications, Beijing, 100876, China.
\IEEEcompsocthanksitem E-mail:\{7597892, nanguo2021, qjwww, wwch, 2024111038, mhq, 2024111039, taoxf\} @bupt.edu.cn. 
\protect\\ \textbullet \quad P. Mao is in Terminus Technologies Co., Ltd. E-mail: mao.pengxuan@tslsmart.com.
\protect\\ \textbullet \quad J. Liu is in School of Computing and Communications, Lancaster University. E-mail: j.liu81@lancaster.ac.uk.
\protect\\ \textbullet \quad Corresponding author: G. Nan (email: nanguo2021@bupt.edu.cn).
}
}

\markboth{Journal of \LaTeX\ Class Files,~Vol.~14, No.~8, August~2021}%
{Shell \MakeLowercase{\textit{et al.}}: A Sample Article Using IEEEtran.cls for IEEE Journals}


\maketitle

\begin{abstract}
Recent advancements in video anomaly understanding (VAU) have opened the door to groundbreaking applications in various fields, such as traffic monitoring and industrial automation. 
While the current benchmarks in VAU predominantly emphasize the detection and localization of anomalies.
Here, we endeavor to delve deeper into the practical aspects of VAU by addressing the essential questions: ``what anomaly occurred?'', ``why did it happen?'', and ``how severe is this abnormal event?''.
In pursuit of these answers, we introduce a comprehensive benchmark for Exploring the Causation of Video Anomalies (ECVA). 
Our benchmark is meticulously designed, with each video accompanied by detailed human annotations.
Specifically, each instance of our ECVA involves three sets of human annotations to indicate ``what'', ``why'' and ``how'' of an anomaly, including 1) anomaly type, start and end times, and event descriptions, 2) natural language explanations for the cause of an anomaly, and 3) free text reflecting the effect of the abnormality.
Building upon this foundation, we propose a novel prompt-based methodology that serves as a baseline for tackling the intricate challenges posed by ECVA.
We utilize ``hard prompt'' to guide the model to focus on the critical parts related to video anomaly segments, and `` soft prompt'' to establish temporal and spatial relationships within these anomaly segments. 
Furthermore, we propose AnomEval, a specialized evaluation metric crafted to align closely with human judgment criteria for ECVA. 
This metric leverages the unique features of the ECVA dataset to provide a more comprehensive and reliable assessment of various video large language models.
We demonstrate the efficacy of our approach through rigorous experimental analysis and delineate possible avenues for further investigation into the comprehension of video anomaly causation.
Our code and dataset are available at \url{https://github.com/Dulpy/ECVA}.
\end{abstract}
\section{Introduction}

\IEEEPARstart{A}nomalies represent occurrences or scenarios that deviate from the norm, defying expectations and straying from routine conditions \cite{9151050, acsintoae2022ubnormal,VideoAnomaly,adam2008robust}. 
These events are typically characterized by their unique, sudden, or infrequent nature, often demanding special attention or intervention \cite{ramachandra2020street}. 
\begin{figure}[t!]
    \centering
    \includegraphics[width=1.0\linewidth]{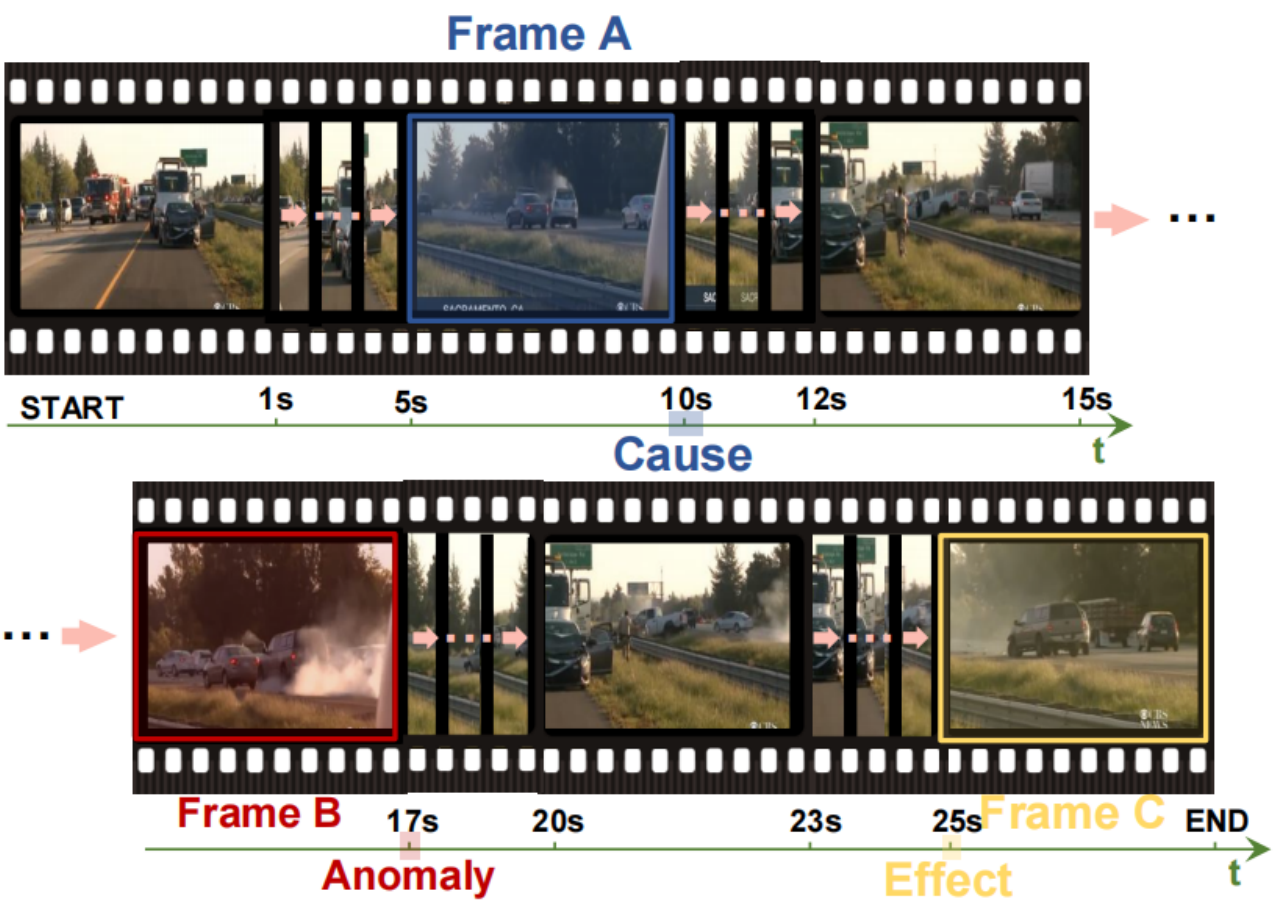}
    \caption{\textbf{Illustration of causations of video anomaly.} The clip started at Frame B refers to a traffic accident, which was caused by the event indicated with Frame A 7 seconds before. The clip in Frame C shows the effect of such an anomaly. A model needs to understand such a long-range relation in the video to yield correct text-based explanations.}
    \label{fig:intro}
    \vspace{-1mm}
\end{figure}
Recently proliferated video anomaly understanding (VAU) \cite{lv2023unbiased,wu2022self} aims at automatically comprehending such abnormal events in videos, thereby facilitating various applications such as traffic surveillance,  environmental monitoring, and industrial manufacturing. 
Towards this direction, video anomaly detection and localization, which refer to identifying abnormal occurrences, and localizing temporally or spatially locate anomalous events in videos, have attracted enormous attention \cite{mehran2009abnormal,wu,velastin2017people,xu2022tad,de2022camnuvem,Wang2023,10083244,Birds,10203301}.

Existing VAU benchmarks \cite{cao2023new,han2022adbench,thakare2023rareanom} and approaches \cite{zaheer2022generative,singh2023eval,chang2022video,nan1,nan2,shi1,shi2,liujun1, liujun2} primarily focus on the aforementioned anomaly detection and localization tasks, while the underlying cause and the corresponding effect of these occurrences, are still largely under-explored. These cues are crucial for perceiving the abnormality and making decisions based on human-interpretable explanations. Figure \ref{fig:intro}
demonstrates a scene of a traffic accident involving many vehicles. ``The accident occurred because a white car parked by the roadside, and a dark gray car traveled at high speed to swerve and rear-end the black car next to it.'' Challenges of comprehending such a cause of the accident include: 1) \textit{capturing key cues in the long video:} a model needs to recognize the white car at the moment indicated by Frame A, which is $7$ seconds before the accident in the clip indicated by Frame B. It is challenging for a model to capture such a long-range relation. 2) \textit{building a logic chain of the cause-effect:} a model needs to further learn rich interactions among clips in the video, indicated by Frame A, Frame B, and Frame C, to build a logic chain of causation of the anomaly, facilitating the generation of the explanations and results. The above two challenges require the development of causation understanding methods that specifically take these characteristics of video anomaly into consideration.

Previous works have demonstrated the great importance of leveraging large, high-quality, and challenging benchmarks to develop and evaluate the state-of-the-art deep learning methods for the VAU task \cite{anomalyde, liu2023generating, aboah2021vision, tian2021weakly, nguyen2019anomaly}. 
Along this line, existing benchmarks have shown their promise \cite{wu, VideoAnomaly, ramachandra2020street, wu2021star_situated_reasoning, liu2024etbench}. Towards VAU in more practical real-world scenarios, they have some limitations:
1) \textit{Lack of cause and effect explanations.} Existing annotations involve the periods when anomalies occur, without providing an explanation of the underlying cause and the effect, as well as the descriptions of targeting anomaly.  
2) \textit{Lack of proper evaluation metrics.} Some traditional evaluation metrics such as BLEU \cite{papineni2002bleu} and ROUGE \cite{lin2004rouge} do not effectively capture semantics, while some GPT-based evaluation metrics suffer from issues of instability and insufficient.
3) \textit{Limited length of videos.} In real-world scenarios, a piece of video may include more than $1.5$ minutes \cite{apostolidis2021video}. However, samples in existing VAU usually have fewer than $30$ seconds, which greatly simplifies the challenges of VAU in real-world cases. 
4) \textit{Limited number of categories.} Anomalies manifest differently across various scenarios. The dataset needs to cover multiple scenarios to ensure the model's generalization and robustness, allowing it to be better applied in real-world.
\begin{figure*}[!ht]
    \vspace{0pt}
    \centering
    \includegraphics[width=1\linewidth]{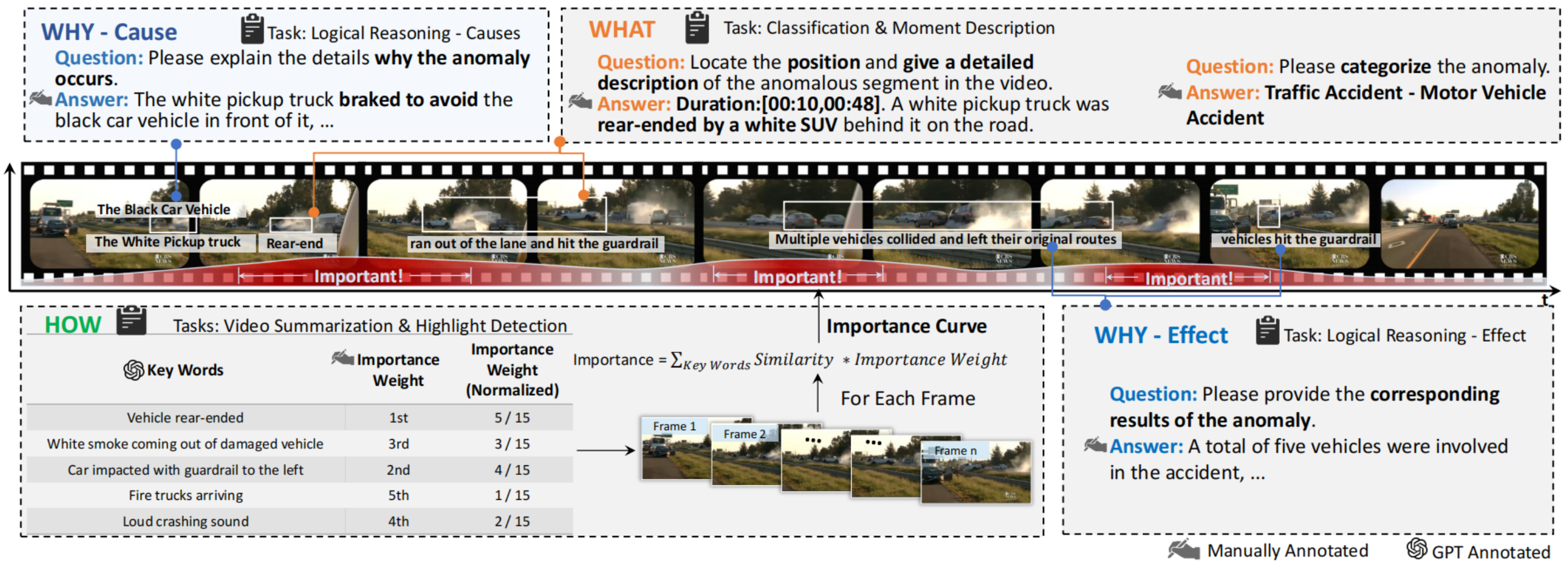}
    \caption{
    \textbf{Overview of the proposed ECVA benchmark.} Our ECVA benchmark consists of manual text-based annotation, including detailed explanations of cause (Why) and effect (Why), anomaly types (What), detailed event descriptions (What), as well as importance scores that can form a curve of events (How).}
    \vspace{0cm}
    \label{fig:dataset}
\end{figure*}

The above limitations of existing datasets call for a benchmark of Causation Understanding of Video Anomaly.
Towards that, we present ECVA, a comprehensive benchmark that contains high-quality annotations of $2,240$ videos from the real world, covering $21$ major categories, and $100$ subcategories of different anomaly types, each involving a 141-second long video and ``621.1'' tokens across ``7.9'' sentences on average.
Specifically, we manually write free-text explanations to detail the underlying cause and the corresponding effects, the descriptions of these events, and the relationships among them. 
To tackle the challenges within ECVA, we introduce a novel prompt-based approach ``AnomSheild'' that harnesses the power of large language models. 
Our method employs the ``hard prompt" to concentrate the model's attention on the anomaly parts of the video.
While the ``soft prompt" facilitates the construction of a coherent and efficient cause-effect logical chain. 
Moreover, we come up with a novel evaluation metric ``AnomEval'' to measure the capability of the methods on the challenging ECVA, which is more comprehensive, robust, and accurate.
Experimental results show the superiority of the proposed method and the evaluation metric. The main contributions of our work can be summarised as follows:

\begin{itemize}
        \item We develop ECVA, a new benchmark for causation understanding of video anomaly. To the best of our knowledge, ECVA is the first large-scale benchmark focused on the causation of video anomalies. Compared with existing datasets, our dataset is more comprehensive and more challenging with much higher-quality annotations.
        \item To tackle the challenges within ECVA, we introduce a novel prompt-based method ``AnomShield''. Specifically, we leverage ``hard prompt'' to capture the key cues of anomaly and  ``soft prompt'' to build a causal logic chain. 
        \item To achieve a more comprehensive, robust, and precise evaluation of the various VLMs' performance on ECVA. We propose a novel evaluation metric ``AnomEval'', which shows high consistency with human's preference.
	\item We conduct extensive experiments on the proposed ECVA. Results show that ECVA enables us to develop and evaluate various VLM methods for causation understanding of video anomalies closer to real-world cases.
\end{itemize}

\section{Related Work}
\label{sec: Related Work}
\begin{table*}[!ht]
\centering
\caption{\textbf{Comparisons between the proposed ECVA and existing VAU datasets.}
Our ECVA is the first large-scale benchmark for causation understanding of video anomaly. It encompasses samples from 100 domains, such as vandalism, traffic accidents, fire incidents, and pedestrian incidents, etc. ECVA sub-tasks primarily focus on the evaluation of causation understanding of video anomaly, and these tasks answer the ``What'', ``Why'' and ``How'' of an anomaly. All textual descriptions or explanations are annotated in \textbf{free-text} format. Here
\textbf{A.C.L.} typically stands for ``Average Clip Length''.
 }
\label{comparison}
\resizebox{1.0\linewidth}{!}{
\begin{tabular}{lcccccccccc}
\toprule
\multirow{2}{*}{Dataset} & \multirow{2}{*}{Domain} & \multicolumn{4}{c}{Video} & \multirow{2}{*}{\# Anomaly Types} & \multicolumn{4}{c}{QA}  \\
\cmidrule(lr){3-6} \cmidrule(lr){8-11}
 &  &  \# Total Frames & Total Length & A.C.L & Audio & & Localization & Description & Reasoning & Outcome  \\
\midrule

UCF-Crimes \cite{sultani2018real} & Crime & 13,741,393 & 128.0h& 242.5s & No & 13 & Frame & NA & NA & NA \\
XD-Violence \cite{wu} & Volence  & 114,096 & 21.07h & 164.3s & Yes & 6 & Frame & NA & NA & NA \\
ShanghaiTech \cite{luo2017revisit}  & Pedestrian & 317,398 & - & - & No & 13 & Bounding-box & NA & NA & NA \\
UCSD Ped1 \cite{wang2010anomaly} & Pedestrian & 14,000 & 0.1h & 6.6s & No & 5 & Bounding-box& NA & NA & NA \\
UCSD Ped2 \cite{wang2010anomaly} & Pedestrian & 4,560 & 0.1h & 6.6s & No & 5 & Bounding-box & NA & NA & NA \\
CUHK Avenue \cite{abnormal2013lu} & Pedestrian & 30,652 & 0.5h & 1.4s & No & 5 & Bounding-box & NA & NA & NA \\
TAD \cite{xu2022tad} & Trafﬁc & 721,280 & 1.2h & 36.8s & Irrelevant & 4 & Bounding-box & NA & NA & NA \\
Street Scene \cite{ramachandra2020street} & Trafﬁc & 203,257 & 380.6s & 3.7s & No & 17 & Bounding-box & NA & NA & NA \\
CamNuvem \cite{de2022camnuvem} & Robbery & 6,151,788 & 57h & 192.2s & No & 1 & Frame & NA & NA & NA \\
Subway Entrance \cite{adam2008robust} & Pedestrian &  86,535 & 1.5h & - & No & 5 & Frame & NA & NA & NA \\
Subway Exit \cite{adam2008robust} & Pedestrian & 38,940 & 1.5h & - & No & 3 & Frame & NA & NA & NA \\
UCF–Crime Extension \cite{ozturk2021adnet} & Crime &  734,400 & 7.5h & 112.5s & No & 1 & Frame & NA & NA & NA \\
BOSS \cite{velastin2017people} & Multiple & 48,624 & 0.5h & 660.0 s & No & 11 & Frame & NA & NA & NA \\
UMN \cite{mehran2009abnormal} & behaviors & 3,855 & 0.1h & 29.1s & No & 1 & Frame & NA & NA & NA \\
UBnormal \cite{acsintoae2022ubnormal} & Multiple & 236,902 & 2.2h & 14.6s & No & 22 & Pixel-level & NA & NA & NA \\
NWPU Campus \cite{Cao_2023_CVPR} & Multiple & 1,466,073 & 16.29h & 107.0s & No & 43 & Frame & NA & NA & NA \\
SUTD-TrafficQA \cite{Xu_2021_CVPR} & Traffic & 1,200,000 & 11h & 4s & No & 1 & Frame & NA & Choice & NA \\
\midrule
CUVA (Ours preliminary version) \cite{Du_2024_CVPR} & Multiple & 3,345,097 & 32.5h & 117.0s & Yes & 42 & Time Duration & Free-text & Free-text & Free-text \\
\textbf{ECVA (Ours)} & Multiple & \textbf{19,042,560} & 88.16h & 141.9s & Yes & \textbf{100} & \textbf{Time Duration} & \textbf{Free-text} & \textbf{Free-text} & \textbf{Free-text} \\ 

\bottomrule
\end{tabular}}

\end{table*}
\noindent
\textbf{Anomaly Datasets:} 
Existing VAU datasets primarily focus on anomaly detection\cite{SceneDe,SingleScene,SwinAnomaly} and localization\cite{VAL,VADL,Singh_2024_CVPR}, and can be broadly categorized into weakly-supervised ones \cite{sultani2018real,wu}, and semi-supervised ones \cite{luo2017revisit,acsintoae2022ubnormal,ramachandra2020street}.
Weakly supervised datasets aim to enhance the model's generalizability, transferring the anomaly detection capabilities from the training set of normal videos to the test set of abnormal videos\cite{abnormal2013lu}.
Semi-supervised datasets emphasize the time points or time periods of anomalous events based on frame level or video-piece level annotations \cite{wang2010anomaly,luo2017revisit}.
Early semi-supervised datasets are mainly designed to support anomaly detection (VAD) and classification tasks\cite{ramachandra2020street,xu2022tad}, with a limited range of anomaly categories and short average video lengths.
Recent VAD datasets\cite{abnormal2013lu,Cao_2023_CVPR} leverage BMP or matrix formats to conduct anomaly localization tasks.
Additionally, some datasets\cite{acsintoae2022ubnormal} simulate real-world anomaly events within virtual environments, enabling pixel-level annotation for a more precise characterization of anomalies.
Our ECVA significantly differs from the existing datasets in these aspects. 
We manually annotate the ``What, Why and How'' of the anomaly, and leverage the importance curve to quantify the intensity of the anomaly.
More detailed comparisons are available in Table \ref{comparison}.


\noindent
\textbf{Methods:} 
The recent strides in multimodal understanding have primarily been driven by the integration of image-based vision models with LLMs.
The majority of video large language models (VLMs) first conduct uniform sampling of video content, followed by the transformation of frames into image tokens through a variety of image encoders such as CLIP\cite{clip} or DINO\cite{zhang2022dino}.
To accommodate the limited context length of large language models, VLMs employ different vision-language adapters (e.g. cross-attention\cite{Ma2022XCLIPEM,zhang-etal-2023-video}, Q-former\cite{Li2023BLIP2BL}, and Pooling operations) to compress these video tokens.
However, this approach falls short when addressing the task of anomaly understanding in long videos within ECVA. 
To address the above issues, we propose a novel method called ``AnomShield''.
Specifically, we leverage ``hard prompt'' to conduct non-uniform sampling of videos to help models concentrate on the anomaly parts of the video. 
Then we develop an efficient mechanism to extract the spatial-temporal relationships within these video parts, which significantly enhances the model's ability to analyze anomaly video content.

\noindent
\textbf{Evaluation Metrics:} 
VAU evaluation metrics \cite{xu2023critical} include,
reference-based ones such as ROUGE \cite{lin2004rouge} and BLEURT \cite{sellam2020bleurt}, answer-based ones such as BLEU \cite{papineni2002bleu}, Rankgen \cite{krishna2022rankgen} and QAFactEval \cite{fabbri2021qafacteval}, and others such as Longformer \cite{beltagy2020longformer}, UniEval \cite{zhong-etal-2022-towards} and MoverScore \cite{zhao-etal-2019-moverscore}.
They essentially evaluate the accuracy and completeness of the generated text by calculating the overlap between the generated text and the reference text in different ways.
However, they still have shortcomings in capturing semantics.
Recently, various GPT-based metrics \cite{xie2023funqa,touchstone,VisITBench} have been developed. 
Compared to traditional methods, GPT-based approaches not only provide a more accurate assessment of generated answers but also consider various dimensions such as richness and detail of the generated answers.
Nevertheless, due to the complexity of anomalies and instability of GPT, these evaluation metrics fail to adequately assess the results.
We propose a novel evaluation metric ``AnomEval'' based on the unique characteristics of the ECVA dataset.
AnomEval not only identifies VLMs' hallucinations but also provides a more comprehensive and robust assessment of the generated answers.

Finally, this work is an extension version from~\cite{Du_2024_CVPR}. The main comparison between our work and the previous one can be summarised as follows:
\begin{itemize}
    \item We expand upon the CUVA dataset in \cite{Du_2024_CVPR}. Specifically, the new dataset, ECVA, contains twice the amount of data, 2.5 times the number of scenarios, and nearly three times the total video duration compared to the CUVA.
    \item To better address the challenges within ECVA, we propose a novel prompt-based method ``AnomShield''. Compared to the previous approach ``A-Guardian'' in \cite{Du_2024_CVPR}, the newly proposed method yields performance gains of $22\%$, $23\%$, and $20\%$ in the cause, description, and effect tasks.
    \item We additionally propose a new VLM evaluation metric for ECVA ``AnomEval'', which is more comprehensive, robust and accurate.
\end{itemize}
\section{The Proposed ECVA Benchmark} \label{sec:Dataset}
In this section, we first introduce our ECVA sub-tasks. 
Then we show how we collect and annotate data.
We also provide a quantitative analysis of the benchmark. 
The overview of our ECVA is demonstrated in Figure \ref{fig:ECVA}.
\subsection{Task Definition}
\begin{figure}[!ht]
    \centering
    \includegraphics[width=0.45\textwidth]{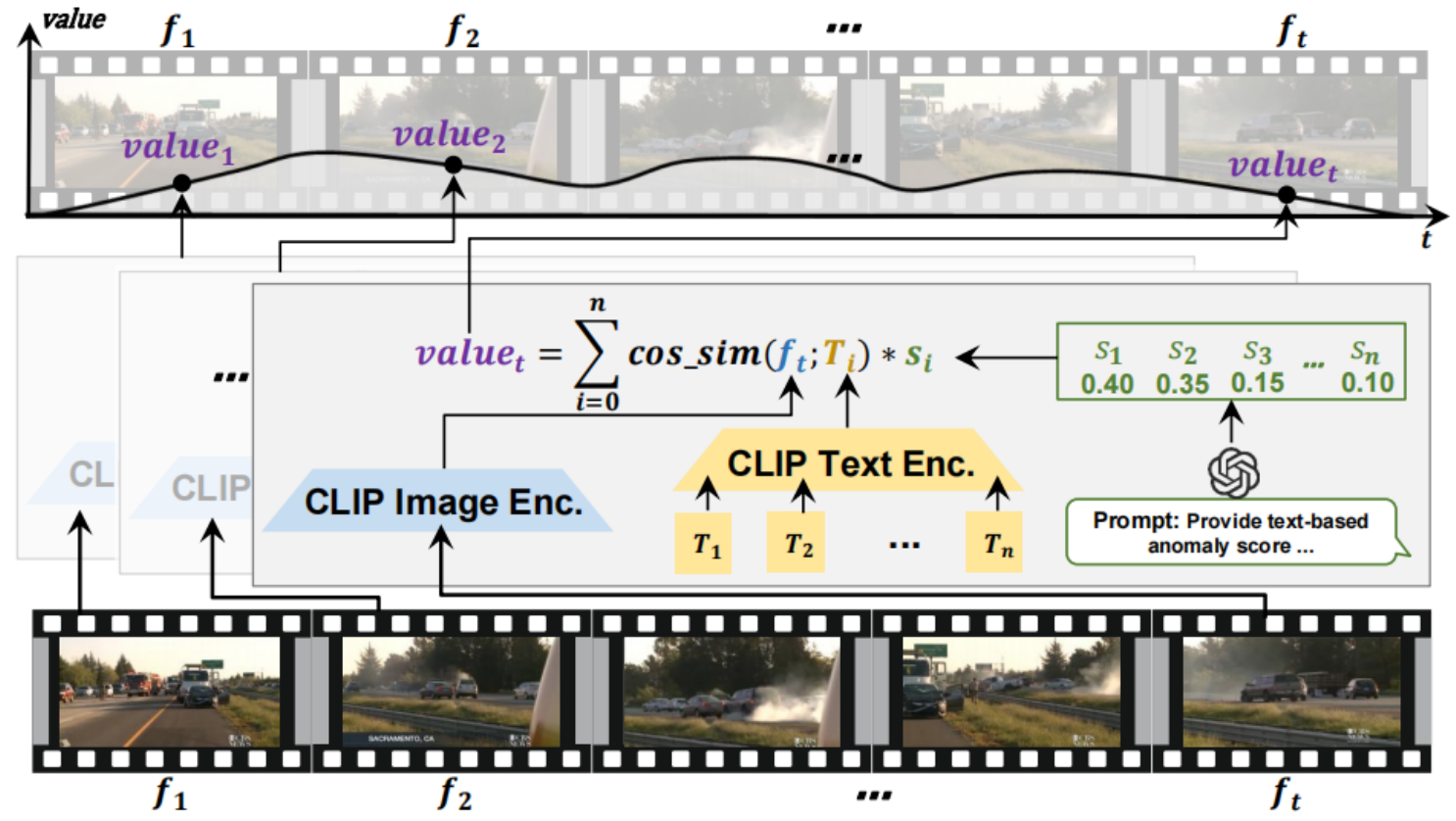}
    \caption{\textbf{Pipeline of generating an importance curve.} Annotators need to consider previous tasks (e.g., Logical Description, Moment Description) and video content to create $3$ to $6$ short sentences ${T_{i}}$ describing all events in the video. We rank these sentences' anomaly severity by Chat GPT \cite{chatgpt} and obtain anomaly scores $s$. Simultaneously, we sample frames ${f_{t}}$ from the video and use CLIP \cite{clip} to measure the similarity between sentences and frames. The resulting similarity scores are multiplied by the anomaly scores for each sentence to get $value_{t}$ for each frame.}
    \label{fig:pipline}
    
\end{figure}

\textbf{What anomaly occurred}: This task includes two objectives: anomaly classification and anomaly description. \textit{Anomaly Classification} includes all the anomaly classes present in the video, which are taken from our database of predefined anomaly classes as shown in Figure \ref{fig:classes}.  
Each video has multiple anomaly classes at different levels, and this task will challenge the model's ability to detect anomaly classes at multiple levels of granularity. 
\textit{Anomaly Moment Description} includes the timestamp in which the anomaly occurs and a detailed description of the anomalous event. 
\newline\textbf{Why this anomaly happened}: This task aims to describe the causal relationships within the video.
Anomaly reasoning describes the reasons for the occurrence of anomalies in the video. 
This task requires the model to infer the cause of the anomaly based on the video content and describe it in natural language, which challenges the model's ability of video comprehension and reasoning.
Anomaly results primarily describe the impacts caused by anomalous events in the video. 
It mainly requests that the model combine external world-knowledge to handle details of anomalous events in the video.
\newline\textbf{How severe this anomaly}: This task aims to reflect the changing trends in the severity of anomalies within the video.
Thus, we propose a novel annotation approach called the importance curve.
Details of our importance curve's pipeline can be found in Figure \ref{fig:pipline}.
The algorithm of the importance curve is illustrated in Algorithm \ref{alg of curve}.
This approach has three advantages: 
1) It provides an intuitive representation of the temporal variation in anomaly severity within the video. 
2) It offers a more intuitive depiction of the inherent causal relationships among anomalous events in the video. 
3) Such an approach enables us to unify various Video Temporal Grounding labels and tasks (e.g. Moment Retrieval, Highlight Detection, Video Summarization) under the same framework.
\begin{algorithm}[ht]
    \renewcommand{\algorithmicrequire}{\textbf{Input:}}
    \renewcommand{\algorithmicensure}{\textbf{Output:}}
\footnotesize
  \caption{Generating importance curve}
    \label{alg:curve}
  {\textbf{Input parameters: video, prompt} } \\
  {\textbf{Output:} {importance curve }} \; 
  \begin{algorithmic}
  \Statex  video\_clip, timestamp = sparse\_sampling(video)
  \Statex  prompt $\rightarrow$ txt\_gpt\_weight
  \Statex  txt\_emb, vid\_emb = Clip(video\_clip, prompt)
  \Statex  txt\_norm\_emb, vid\_norm\_emb = normalization(txt\_emb, vid\_emb)
  \Statex  similarity = cosine\_similarity(txt\_norm\_emb,vid\_norm\_emb)
  \Statex  similiar\_score = normalization(similarity) 
  \Statex  imp\_value = similiar\_score * txt\_gpt\_weight
  \Statex  timestamp, imp\_value $\rightarrow$ importance curve
  \end{algorithmic}
  \label{alg of curve}
\end{algorithm}

\subsection{Dataset Collection}
We crawled data from prominent video platforms such as Bilibili and YouTube\footnote{We have obtained permission from Bilibili \url{www.bilibili.com} and YouTube \url{www.youtube.com} to use their video data for non-commercial purposes.}. 
We discarded videos that encompass sensitive themes such as pornography and politics. 
Throughout the data collection process, we thoroughly analyze the quantity and quality of videos in each category, which in turn lead to the selection of the final $2500$ anomalous videos. 
These videos are then categorized into $21$ main categories, such as ``robbery", ``traffic accident" and ``fire".  
Each major category is further divided into subcategories. 
For example,  we divided the ``fire'' category into the ``commercial building fire'', ``forest fire'', ``factory fire'' and ``residential fire'' subcategories. 
In this way, we obtain $100$ subcategories in total.  
\subsection{Annotation Pipeline}
\label{section 3.3 ano pip}
Our dataset construction pipeline involves three stages: pre-processing, manual annotation, and importance curve processing. 
The whole process takes about $300$ hours with over $25$ annotators.

\subsubsection{Pre-processing}
First, we develop a script to crawl a huge number of videos from Bilibili and YouTube. 
To ensure these videos meet our project standards for resolution, clarity, and relevance, we try to improve video quality by deleting blurring parts and filter out the irrelevant video. 
Then we conduct a thorough manual screening to exclude videos that does not adhere to ethical standards or contains sensitive information. 
Throughout the dataset collection and annotation process, we strictly follow the ethical requirement of the website. 
Finally, $2,240$ anomaly video clips are obtained.

\begin{figure*}[ht]
   \centering
    \begin{minipage}{0.5\textwidth}
        \centering
        \subfloat[]{
            \includegraphics[width=1.0\linewidth]{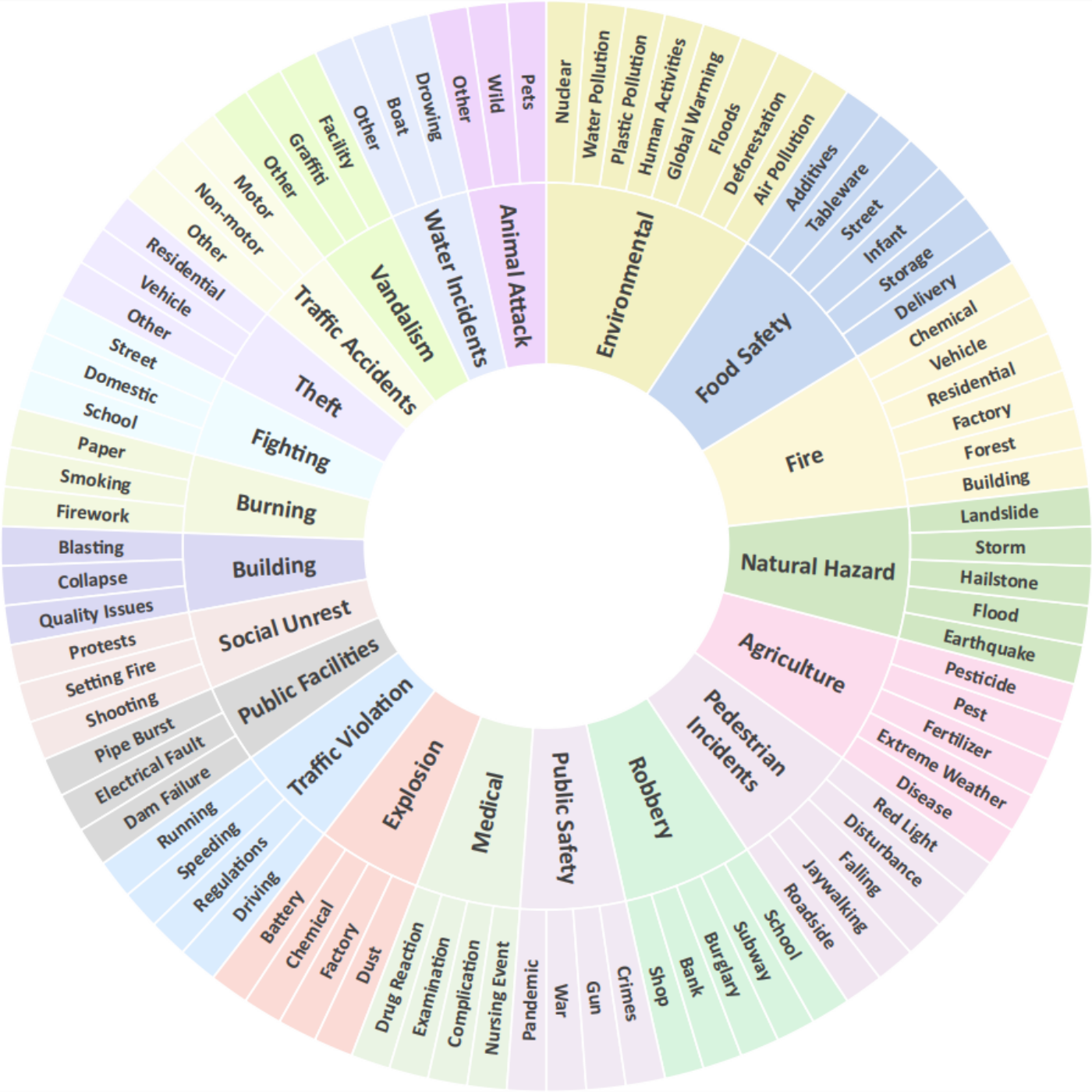}
            \label{fig:classes}
        }
    \end{minipage}%
    \begin{minipage}{0.5\textwidth}
        \centering
        \subfloat[]{
            \includegraphics[width=1.0\linewidth]{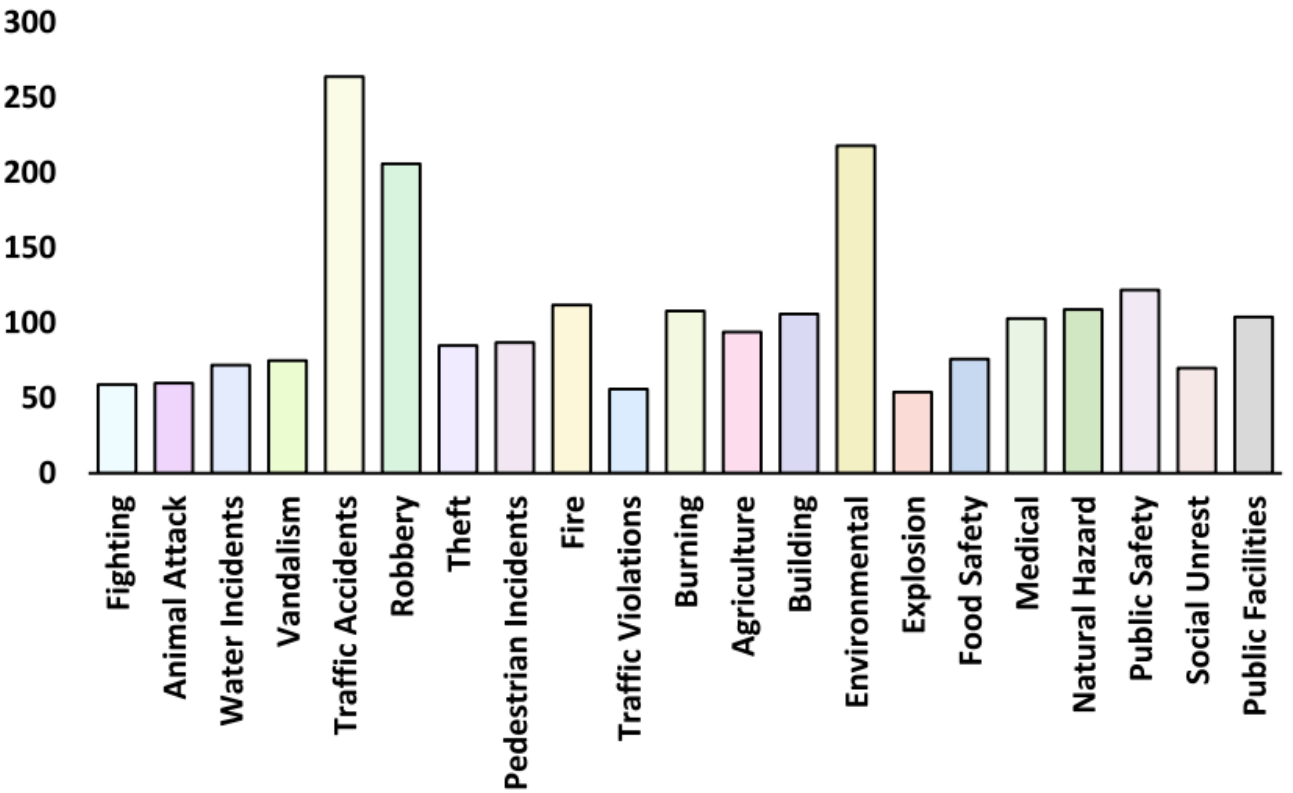}
            \label{fig:class-count}
        }
    \end{minipage}

    \begin{minipage}{0.33\textwidth}
        \centering
        \subfloat[]{
            \includegraphics[width=1.0\linewidth]{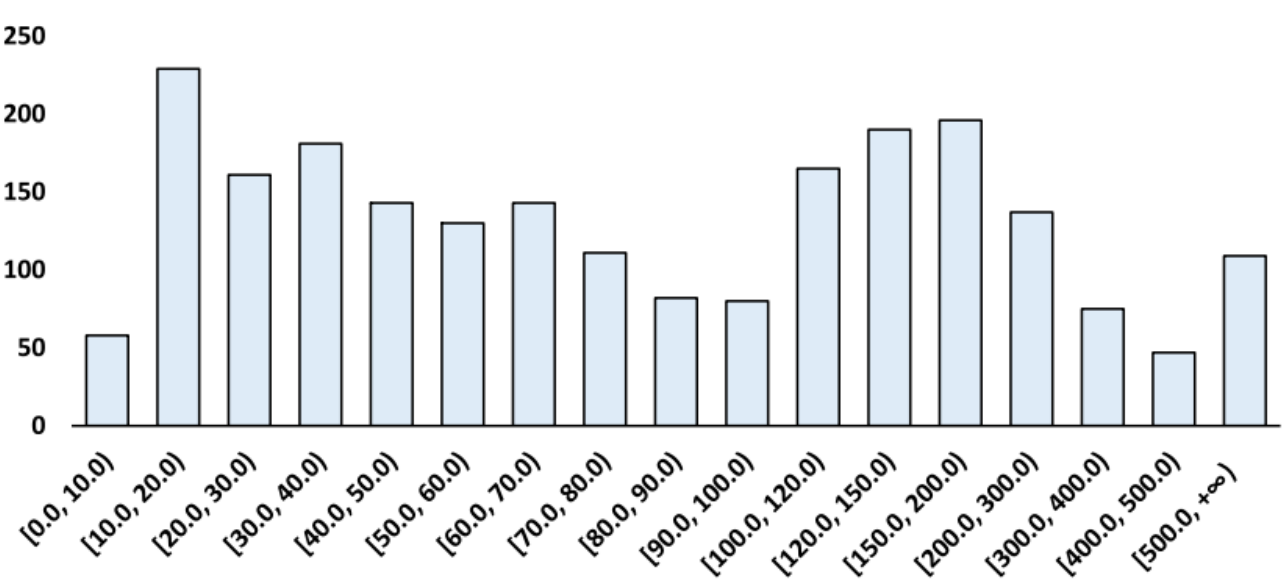}
            \label{fig:video-time}
        }
    \end{minipage}%
    \begin{minipage}{0.33\textwidth}
        \centering
        \subfloat[]{
            \includegraphics[width=0.8\linewidth]{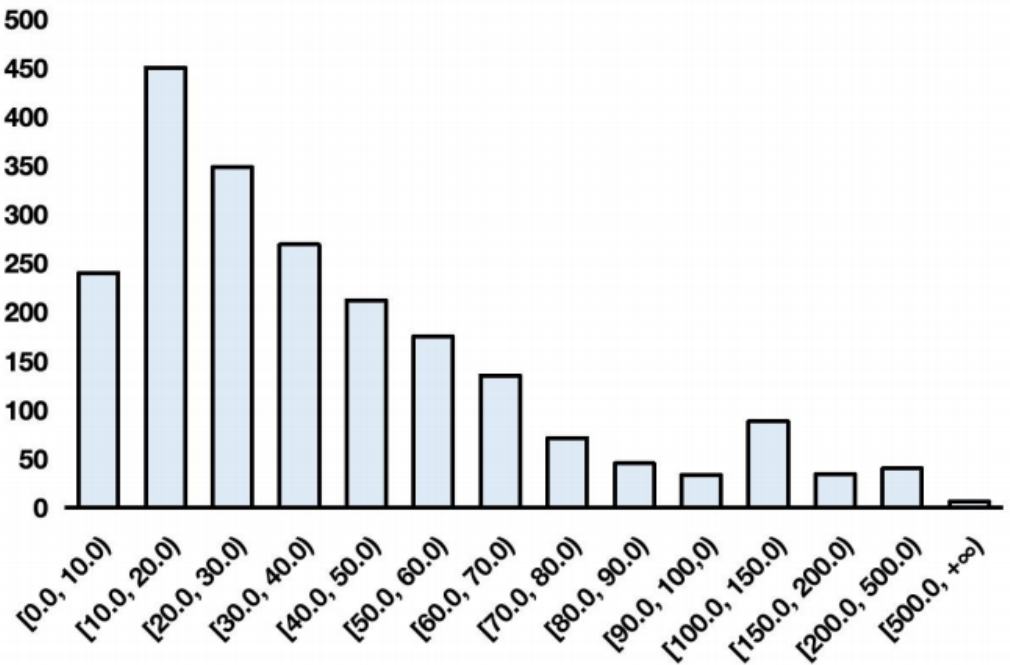}
            \label{fig:anomaly-time}
        }
    \end{minipage}%
    \begin{minipage}{0.33\textwidth}
        \centering
        \subfloat[]{
            \includegraphics[width=0.8\linewidth]{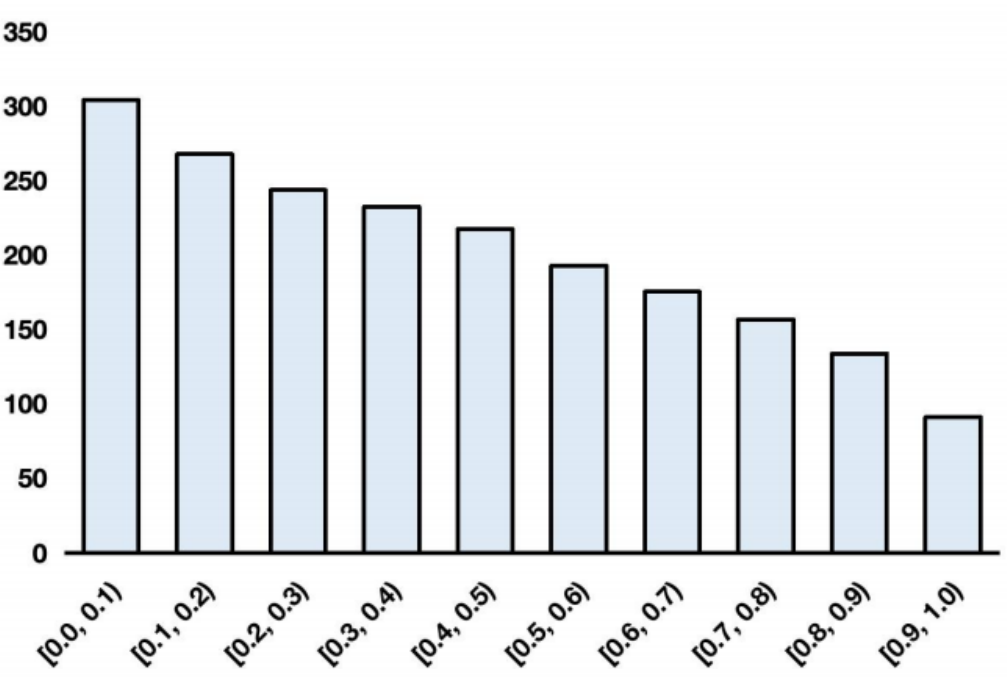}
            \label{fig:per}
        }
    \end{minipage}
    \vspace{0cm}
    \caption{\textbf{Statistics of our ECVA dataset.} Figure (a) shows all anomaly types in ECVA. Figure (b) shows the number of videos in each anomaly type. Figure (c) shows the distribution of video length. Figure (d) shows the distribution of anomaly segment duration. Figure (e) shows the temporal distribution of anomalous segments. }
\label{fig:ECVA}
\end{figure*}

\subsubsection{Manual Annotation}
We annotate the videos in English according to the designed annotation document, and we split the annotation procedure into two rounds.
We employ a mechanism similar to kappa \cite{XIA2020309} to screen and train annotators, ensuring the consistency of their annotation content.
In the first round, we ask annotators to annotate all videos according to the task definition. 
Then, in the second round, we shuffle the annotators and ask these annotators to review and supplement the annotation results of the first round. 
This approach not only allows for the refinement and enhancement of the video descriptions but also significantly improves the overall accuracy and reliability of the annotated data.
\subsubsection{Post-processing of Importance Curve}
Due to the limited capabilities of the CLIP model and sampling intervals, the initial curve may fail to accurately reflect the time periods of anomalies, which significantly impacts the effectiveness of downstream tasks. 
Thus, we incorporate the following three tasks to optimize the importance curve, such as Video Captioning \cite{2023videochat}, Video Entailment \cite{sevila}, and Video Grounding \cite{univtg} respectively. 
Based on these tasks, we employ a voting mechanism to precisely identify the time segments in the video corresponding to the given key sentences.
\subsection{Dataset Statistics}
Our ECVA dataset contains $2,240$ video clips and $6,720$ question-answer pairs, the total length of these videos is $88.16$ hours, and the average frames of videos is $8,460$. 
The frames are extracted from the original videos at a rate of $60$ FPS.
The videos encompass a wide range of domains. 
Finally, we categorize anomaly events into $21$ scenarios, resulting in a total of $100$ types of anomalies, as illustrated in Figure \ref{fig:ECVA}(a). 
The distribution of video categories and length are illustrated in Figure \ref{fig:ECVA}(b) and \ref{fig:ECVA}(c), respectively.
The distribution of anomaly segment duration can be found in Figure \ref{fig:ECVA}(d), and the percentage of video time proportions shown in Figure \ref{fig:ECVA}(e). 
\section{The Proposed Method: Anomaly Shield}
\begin{figure*}[!ht]
    \centering
    \includegraphics[width=1.0\textwidth]{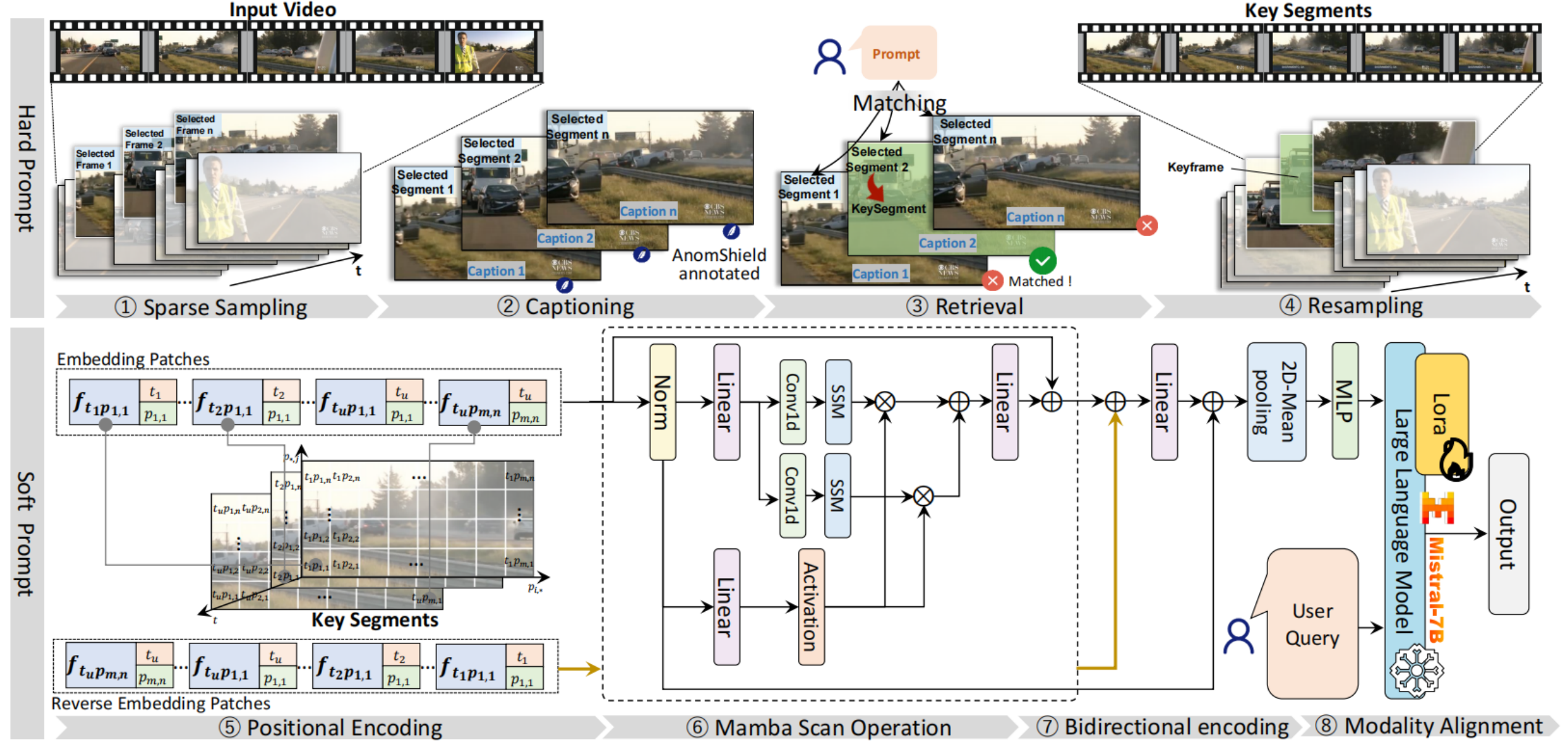}
    \caption{\textbf{The architecture of our AnomShield.} We first conduct sparse uniform sampling for each video\normalsize{\textcircled{\scriptsize{1}}}\normalsize, and then apply AnomShield (optimized through a two-stage training process) to generate descriptions for the sampled frames\normalsize{\textcircled{\scriptsize{2}}}\normalsize. Next, we identify key frames in the video by a matching strategy\normalsize{\textcircled{\scriptsize{3}}}\normalsize and conduct dense sampling around these key frames to capture the essential segments of the video\normalsize{\textcircled{\scriptsize{4}}}\normalsize. For these key segments, we add spatial-temporal position embedding to each frame\normalsize{\textcircled{\scriptsize{5}}}\normalsize and leverage a bidirectional Mamba-based method to extract their spatio-temporal relationship\normalsize{\textcircled{\scriptsize{6}}}\normalsize \normalsize{\textcircled{\scriptsize{7}}}\normalsize. Then, we use an MLP to align text and image features\normalsize{\textcircled{\scriptsize{8}}}\normalsize, and finally feed text-image feature into the base LLM to get the answer.}
    \label{fig:model_arch} 
\end{figure*}
In this section, we introduce a novel video large language model named Anomaly Shield (AnomShield), which is designed to address the two challenges presented by our dataset.
\textit{To effectively capture crucial cues within long videos}, we propose to leverage chain-of-thought prompting to guide VLMs to concentrate on pivotal clues in the video pertinent to the provided questions.
By harnessing the exceptional logical reasoning capabilities of large language models (LLMs) to \textit{build a logic chain of the cause-effect}, we design an efficient pretraining framework to facilitate video-language alignment.

\subsection{Chain-of-Thought Prompting Design}
Traditional VLMs adopt a uniform sampling strategy to capture the information of the video modality. 
However, it is evident that not all frames hold equal significance for understanding anomalies.
Given the relatively long duration of videos within the ECVA and the limited number of video frames that the VLMs can process.
Accurately selecting the anomaly-relevant parts of the video is crucial for the VLM to comprehend the abnormal events precisely.
Our chain-of-thought prompting includes four steps: Sparse Sampling, Captioning, Retrieval, and Resampling.\\
\textbf{Coarse Sampling:} We initially conduct uniform sampling to extract $32$ frames from every piece of video and divide these $32$ frames into $8$ segments.\\
\textbf{Captioning:} We utilize the image understanding capabilities of our pre-trained VLM to generate captions for each segment.\\
\textbf{Retrieval:} We utilize the GPT to meticulously compare the captions of each segment with the user's query. In this way, we are able to accurately pinpoint the positions of the several top key segments in the video that exhibit the highest degree of alignment with the user's query.\\
\textbf{Resampling:} In order to capture the nuances details associated with the anomaly events, we perform dense sampling on these key segments, averaging $8$ frames per segment.
\subsection{Model Structure}
We employ the chain-of-thought prompting technique to selectively sample key frames from the video. 
To capture the nuanced interactions across various levels of visual granularity, each frame is meticulously split into $M$ patches. 
These patches are then processed through an image encoder to distill fine-grained features.
Specifically, we use the CLIP-L \cite{clip} to extract patch-level features denoted as \(\mathbf{P} = \{p^1, p^2, ...,p^m \}\), where \(p^m \in \mathbb{R}^{T \times M \times D}\) and \(D\) is the dimension of each patch-level feature. 
To retain the temporal and spatial position information, we add a trainable temporal position embedding \(p_t \in \mathbb{R}^{T \times D}\) and spatial position embedding \(p_s \in \mathbb{R}^{M \times T \times D}\) for each patch.
\begin{align}
    F = \mathbf{P} + p_t + p_s
\end{align}
Following the spatial-first strategy \cite{vit,ViViT}, we sequence the features of each patch before feeding them into a connector module, which is denoted as \(F_l\).
This connector is designed to effectively extract dynamic spatio-temporal features between patches, and adeptly align the visual features with the word embedding space of Large Language Models (LLMs).
To strike a balance between compute efficiency and the connector's targets, we develop a method based on mamba\cite{mamba2} to extract the video spatial-temporal features.
We first normalize the input token sequence \(F_l\) to obtain a normalized sequence \({F_l}'\) as follows, and the normalization function is denoted as \( \mathcal{N} \).
\begin{align}
    F'_{l} = \mathcal{N}(F_{l})
\end{align}
Feature vectors \( x \) and \( z \) are extracted from the normalized sequence \( F'_{l} \) using linear transformations represented by weight matrices \( \mathbf{W}_x \) and \( \mathbf{W}_z \), respectively:
\begin{align}
    x = \mathbf{W}_x F'_{l} \quad \quad z = \mathbf{W}_z F'_{l}
\end{align}
Following mamba \cite{mamba} architecture, we apply a 1-D convolution operation with an activate function to get  \( x_o' \) and compute the matrices \( B_o \) and \( C_o \) using linear transformations with weight matrices \( \mathbf{W}_{B_o} \) and \( \mathbf{W}_{C_o} \), respectively
\begin{align}
     x_o' = \sigma(\mathbf{K}_o * x) \\
     B_o = \mathbf{W}_{B_o} x_o' \\
     C_o = \mathbf{W}_{C_o} x_o' 
\end{align}
And the \( \Delta_{o} \) is calculated as following:
\begin{align}
\Delta_o = \log(\exp(\mathbf{W}_{\Delta_o} x_o' - \delta) + 1)
\end{align}
We then leverage \( \Delta_{o} \) to obtain \(\overline{A}_{o} \), \(\overline{B}_{o} \).
\begin{align}
\overline{A}_{o} = \Delta_{o} \otimes \text{Parameter}_{o}^{A} \quad \overline{B}_{o} = \Delta_{o} \otimes \text{Parameter}_{o}^{A}
\end{align}
And the token length \( y \) is computed through the S4 model \cite{mamba2}.
\begin{align}
    y = \operatorname{S4}\left(\overline{A}_{o}, \overline{B}_{o}, C_{o}\right)\left(x_{o}^{\prime}\right) \circ \sigma(z)
\end{align}
The final token sequence \( y_l \) is obtained by combining the feature vectors with a residual connection using a weight matrix \( \mathbf{W}_T \) and adding the original sequence.
Furthermore, in order to capture the bidirectional semantic relationships along image patch sequences, we process the feature vectors with both forward \( y' \) and backward \( y \) directions.
\begin{align}
 y_l = \mathbf{W}_T (y' + y) + F_{l} 
\end{align}
We perform the pooling operation to downsample the spatial-temporal feature to meet the input length requirements of LLMs.
Subsequently, we utilize MLP to change the dimension of spatio-temporal features to obtain the final video features.
Finally, we concat the query and video tokens and feed them into the LLM to get the answer.


\subsection{Training Strategy}
In this section, we detail the training process for AnomSheild. We split our training strategy into three stages.\\
\textbf{Stage one:} This stage aims to align the visual features into the word embedding space of LLMs.
Thus, we categorize the image-text data into long-text type and short-text type based on the length of the text.
We freeze the Base LLM and visual encoder while pre-training the connector with the low-quality short-text data, such as CC3M\cite{CC3M}.\\
\textbf{Stage two:} Based on the first stage, we further train the connector using high-quality long-text data, such as shareGPT4V\cite{sharegpt4v}, shareGPT4o\cite{sharegpt4video}, enabling the model to accommodate long-sequence inputs and integrate new information.\\
\textbf{Stage three:} In this stage, we unfreeze the visual encoder and employ the LoRA (Low-Rank Adaptation) to fine-tune the LLM.
We incorporate a variety of video datasets to train the model, including VideoGPT-plus\cite{VideoGPT+} and VideoChat100K\cite{video-chatgpt}.
We further incorporate SUTD\cite{Xu_2021_CVPR} into our training dataset, aiming to enhance the model's ability to understand video anomaly understanding.
\section{Evaluation Metric} \label{sec:Evaluation Metric}
\begin{figure*}[!ht]
    \centering
    \includegraphics[width=1\linewidth]{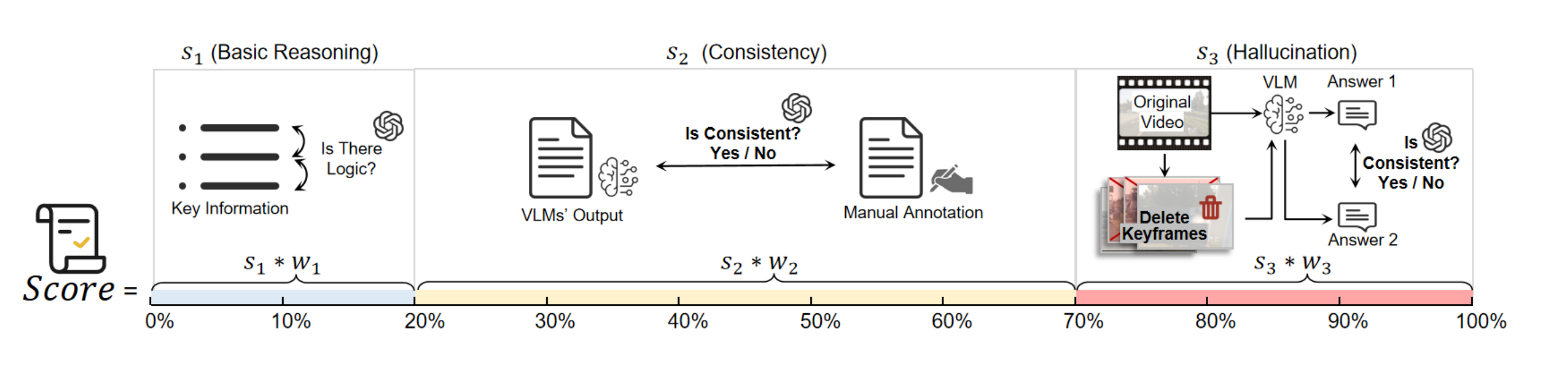}
    \caption{
    \textbf{Overview of the AnomEval.} AnomEval is composed of three key components: Basic Reasoning, Hallucination, and Consistency. In the Basic Reasoning part, we leverage GPT to assess whether the candidate answers comprehensively cover all key phrases and rate the answers based on their logical coherence. For the Consistency evaluation, we leverage the binarity of the GPT to score the candidate answers. As for the Hallucination part, we remove key frames from the video and input the edited videos into the model under evaluation to observe whether the consistency of the model's responses is maintained with or without the key frames. If the model's responses remain highly consistent after the key frames are removed, it indicates that the model is not responding based on the video content.}
    \label{fig:Evaluation_metric}
\end{figure*}
Currently, VLMs widely adopt the GPT-based approach to compare candidate answers with reference answers in terms of coherence, consistency, fluency, and correctness. 
However, this approach has significant limitations:
\begin{itemize}
	\item The multifactorial of causality in anomalous events. Due to the influence of analysis methods and the observer's experience, the causality of the same anomalous event can have multiple interpretations. 
	\item The instability of GPT. For the same sample, GPT may generate completely different evaluations.
	\item Due to the hallucination issue within VLM, the model may provide accurate answers based on training data without truly understanding the content of the video\cite{Hallucinations}.
\end{itemize}

To address the above issues, we propose a stable, comprehensive, and accurate evaluation metric to assess the ability of VLMs to understand anomalous events. As shown in Figure \ref{fig:Evaluation_metric}, our evaluation matrix consists of three core components:\\
\textbf{Basic Reasoning:} To address the multifactorial of causality in anomalous events, we employ GPT to evaluate the quality of candidate answers from two dimensions based on key entities within ECVA: First, we utilize GPT to assess the number of annotated key entities covered by the candidate answers, evaluating how many key entities in the video are covered by the model's responses. Second, we employ GPT to ascertain the rationality of the logic formed by these entities in the candidate answers, which is denoted by equation $5$.
\( K \) represents the set of key sentences for the \( i \)-th sample, \( x \) represents the candidate answer for the \( i \)-th sample, \( G \) denotes the scoring mechanism of GPT, and \( N \) represents the total number of samples.
\begin{equation}
s_1 = \sum_{i=1}^N(\frac{G{(x_i \cap K_i)}}{K_i} * G(x_i))
\end{equation}
\textbf{Consistency:} Given that GPT models exhibit a high level of stability when handling affirmative or negative questions\cite{li2023evaluatingobjecthallucinationlarge}, we leverage this advantage to measure the consistency level between the candidate answer and the reference answer.
This approach significantly enhances the stability of AnomEval. 
As denoted in equation $6$, \( q_t \) denotes the \( t \) question for the \( i \)-th sample, \( y_i \) denotes the ground truth for the \( i \)-th sample, and \( T \) represents the total number of questions.
\begin{equation}
s_2 = \frac{\sum_{i=1}^N\sum_{t=1}^T{G(``Yes"|x_i,y_i,q_t)}}{\sum_{i=1}^N\sum_{t=1}^T{G(``Yes"|x_i,y_i,q_t) + G(``No"|x_i,y_i,q_t)}}
\end{equation}
\textbf{Hallucination:} Since the VLM may provide relatively correct answers based on hallucinations\cite{huang2024opera}. To ensure that VLMs truly understand the content of anomalous videos rather than merely relying on past training data to respond, we remove the anomaly parts in the video and ask the VLM to answer the same questions based on the edited videos. 
We utilize GPT to compare the consistency of the answers before and after the removal. 
A higher consistency score indicates that the VLM has a weaker understanding ability of the video. 
As denoted in equation $7$, \( {x_i}^- \) denotes the answer of VLM based on the edited video.
\begin{equation}
s_3 = -G({x_i}^-,x_i)
\end{equation}
Finally, we assign different weights \( w \) to the scores obtained from each component and sum them up to derive the final score.
\begin{equation}
S = w_1*s_1 + w_2 *s_2 + w_3 *s_3
\end{equation}
In this way, AnomEval can not only comprehensively assess the VLMs' ability to recognize anomalous events but also verify their capability to truly understand video anomaly, thus providing more reliable and accurate evaluation results.
\section{Experiments} \label{sec:Experiment}

\subsection{Implementation Details}
In our study, we employ Clip-Vit-L and Mistral as the visual encoder and Large Language Model (LLM).
All training procedures are conducted on $4$ NVIDIA A800 GPUs. 
We pre-train AnomShield for three epochs with a batch size of $32$, employing the AdamW optimizer with a cosine schedule. 
Specifically, we pre-train our AnomShield with the initial phase requiring $16$ hours, the second phase $10$ hours, and the final phase $20$ hours. 
During the third phase, we sample an average of sixteen frames per video and set the learning rate of the visual encoder to be $20\%$ of that used for the LLM. More details can be found at our github: \url{https://github.com/Dulpy/ECVA}.
\begin{table}[htbp]
\vspace{-5mm}
\centering
\caption{\textbf{Human consistency evaluation}}
\resizebox{0.8\linewidth}{!}{
\begin{tabular}{lccc}
\toprule
\multirow{2}{*}{Metrics} & \multicolumn{3}{c}{Answer Pool Ranking} \\
\cmidrule(l){2-4}
& Description& Cause & Effect  \\
\midrule
ROUGE \cite{lin2004rouge}& 11.1\% & 5.7\% & 10.2\%  \\
BLEU \cite{papineni2002bleu}  & 9.2\% & 3.4\% & 13.1\% \\
BLEURT \cite{sellam2020bleurt} & 15.3\%& 15.6\% & 22.4\%  \\
MoverScore \cite{zhao-etal-2019-moverscore}& 40.1\%& 45.9\% & 38.7\%  \\
UniEval \cite{zhong-etal-2022-towards} & 42.3\%& 42.9\% & 50.4\%  \\
GPT-Base \cite{zhong-etal-2022-towards} & 71.3\%& 73.2\% & 69.7\%  \\
\midrule
\textbf{AnomEval (Ours)}& \textbf{82.3\%} & \textbf{80.2\%} & \textbf{89.1\%}  \\
\bottomrule
\end{tabular}
}
\centering

\label{T:evaluation_zoo}
\end{table}
\begin{table*}[htbp]
\centering
\caption{Here we test \textbf{Cause} task on ECVA benchmark using multiple VLMs and the proposed \textbf{AnomShield}. We conduct evaluations on traditional metrics and our \textbf{AnomEval}. It can be seen that AnomShield achieves the \textbf{second place} on AnomEval.}
\resizebox{1.0\linewidth}{!}{
\begin{tabular}{l|c|c|c|c|c|c|c|c|c}
\toprule
\textbf{Method} & BaseLLM &\textbf{BLEU} & \textbf{ROUGE-2} & \textbf{ROUGE-L} & \textbf{BLEURT} & \textbf{MoverScore} & \textbf{UniEval} & \textbf{GPT-Based} & \textbf{AnomEval} \\
\midrule
mPLUG-owl\cite{mplug-owl} & BLoomZ-7B &0.32\% & 1.69\% & 11.58\% & 31.93\% & 43.46\% & 56.71\% & 3.29 & 17.15\% \\
\midrule
Otter\cite{li2023otter} & \multirow{5}{*}{LLaMA-1-7B} & 0.98\%  & 1.63\% & 13.33\% & 29.30\% & 43.48\% & 51.27\% & 3.04 & 19.32\% \\
PandaGPT\cite{su2023pandagpt} & &0.17\% & 2.18\% & 11.45\% & 44.95\% & 43.44\% & 57.89\% & 3.74 & 24.41\% \\
ST-LLM\cite{ST-LLM} & &0.63\% & 0.56\% & 9.99\% & 29.98\% & 43.48\% & 43.58\% & 2.33 & 12.52\% \\
Video-ChatGPT\cite{video-chatgpt} & &0.14\% & 2.11\% & 9.61\% & 41.20\% & 43.44\% & 48.94\% & 4.30 & 9.93\% \\
A-Guardian\cite{Du_2024_CVPR} & &0.41\% & 2.37\% & 11.15\% & 41.07\% & 43.43\% & 48.64\% & 4.35 & 11.27\% \\
\midrule
TimeChat\cite{Ren2023TimeChat}& \multirow{5}{*}{LLaMA-2-7B}  & 0.85\%  & 2.25\% & 13.15\% & 32.99\% & 43.44\% & 44.14\% & 3.54 & 20.39\% \\
Video-llava\cite{videollava} & & 0.31\% & 2.40\% & 11.50\% & 37.83\% & 43.44\% & 48.99\% & 4.74 & 18.64\% \\
LLaMA-VID\cite{llamavid} & & 0.22\% & 2.55\% & 11.26\% & 38.03\% & 43.45\% & 53.76\% & 5.35 & 24.32\% \\
MovieLLM\cite{song2024moviellm} & & 0.19\% & 0.25\% & 3.73\% & 24.70\% & 43.37\% & 40.92\% & 1.51 & 19.72\% \\
Chat-Univi\cite{Chat-UniVi} & & 0.55\% & 5.41\% & 18.42\% & 37.82\% & 43.46\% & 56.19\% & 6.33 & 19.33\% \\
\midrule
VILA\cite{vila}& LLaMA-3-8B & 0.25\% & 3.45\% & 12.10\% & 38.97\% & 43.46\% & 55.80\% & 5.51 & 35.00\% \\
\midrule
Video-LLaMA2\cite{videollama2}& \multirow{2}{*}{Mistral-7B}  & 0.12\% & 1.93\% & 7.59\% & 42.24\% & 43.44\% & 49.88\% & 5.84 & 22.26\% \\
VideoChat2-mistral\cite{Videochat2}& & 0.43\% & 2.25\% & 12.12\% & 30.20\% & 43.46\% & 60.82\% & 5.06 & 32.70\% \\
\midrule
AnomShield (Ours)& Mistral-7B & 0.79\% & 2.99\% & 14.58\% & 37.03\% & 43.43\% & 47.98\% & 4.82 & 33.30\% \\
\bottomrule
\end{tabular}
}
\label{cause-exp}
\end{table*}
\subsection{Consistency evaluation of AnomEval}
\textbf{Our AnomEval metric can better align with human's preference on causation understanding of video anomaly.} To validate the consistency of our evaluation metric with human judgment, we conduct human consistency experiments. 
Due to the difficulty human experts face in assigning precise scores to specific answers, yet their ease in discerning the quality of answers, we adopt a ranking method to measure the consistency of each evaluation metric. 
Specifically, we extract $1000$ samples from the experimental results, each consisting of a video, a question, and three answers(e.g. second round annotation, first round annotation, GPT generate) based on that video to the question. 
We ask human experts to rank these three answers based on their correctness, completeness and contextual understanding.
The ranking results will serve as the ground truth for Human consistency evaluation.
Meanwhile, we utilize our AnomEval and other evaluation metrics to score these three answers and rank them based on their scores.
Finally, we compare the similarity between these ranking results and ground truth to assess their consistency.
As shown in Table \ref{T:evaluation_zoo}, AnomEval can better align with human preferences.

\begin{table*}[h]
\centering
\caption{Here we test \textbf{Description} task on ECVA benchmark using multiple VLMs and our \textbf{AnomShield}. We conduct evaluations on traditional metrics and our \textbf{AnomEval}. It can be seen that AnomShield achieves the \textbf{first place} on AnomEval.}
\resizebox{1.0\linewidth}{!}{
\begin{tabular}{l|c|c|c|c|c|c|c|c|c}
\toprule
\textbf{Method} & BaseLLM &\textbf{BLEU} & \textbf{ROUGE-2} & \textbf{ROUGE-L} & \textbf{BLEURT} & \textbf{MoverScore} & \textbf{UniEval} & \textbf{GPT-Based} & \textbf{AnomEval} \\
\midrule
mPLUG-owl\cite{mplug-owl} & BLoomZ-7B &0.06\% & 1.04\% & 7.05\% & 22.72\% & 46.49\% & 57.15\% & 3.87 & 20.29\% \\
\midrule
PandaGPT\cite{su2023pandagpt} & \multirow{5}{*}{LLaMA-1-7B} &0.06\% & 1.49\% & 7.66\% & 30.43\% & 46.48\% & 55.76\% & 4.50 & 31.61\% \\
ST-LLM\cite{ST-LLM} & &0.16\% & 0.64\% & 7.64\% & 24.07\% & 46.52\% & 56.00\% & 5.67 & 26.12\% \\
Otter\cite{li2023otter} & &0.51\% & 1.92\% & 11.83\% & 23.82\% & 43.42\% & 54.20\% & 4.99 & 24.30\% \\
Video-ChatGPT\cite{video-chatgpt} & &0.09\% & 1.52\% & 7.62\% & 30.56\% & 46.46\% & 53.77\% & 4.90 & 14.30\% \\
A-Guardian\cite{Du_2024_CVPR} & &0.08\%& 1.45\% & 7.32\% & 32.07\% & 46.46\% & 52.64\% & 4.93 & 17.27\% \\
\midrule
TimeChat\cite{Ren2023TimeChat}& \multirow{5}{*}{LLaMA-2-7B} & 0.41\% & 2.00\% & 11.10\% & 26.55\% & 46.49\% & 54.26\% & 6.02 & 32.83\% \\
Video-llava\cite{videollava}& & 0.03\% & 1.15\% & 5.95\% & 32.41\% & 46.47\% & 53.80\% & 5.89 & 23.02\% \\
LLaMA-VID\cite{llamavid}& & 0.06\% & 1.42\% & 7.25\% & 27.08\% & 46.47\% & 53.18\% & 6.15 & 25.13\% \\
MovieLLM\cite{song2024moviellm}& & 0.03\% & 1.28\% & 6.37\% & 31.12\% & 46.46\% & 51.47\% & 6.30 & 39.46\% \\
Chat-Univi\cite{Chat-UniVi}& & 0.55\% & 5.41\% & 18.42\% & 27.09\% & 46.49\% & 54.10\% & 6.59 & 27.49\% \\
\midrule
VILA\cite{vila}& LLaMA-3-8B & 0.11\% & 1.56\% & 7.83\% & 26.64\% & 46.48\% & 50.25\% & 5.98 & 37.49\% \\
\midrule
Video-LLaMA2\cite{videollama2} & \multirow{2}{*}{Mistral-7B}&0.05\% & 1.30\% & 7.26\% & 26.62\% & 46.49\% & 61.45\% & 6.67 & 28.16\% \\
VideoChat2-mistral\cite{Videochat2}& & 0.13\% & 0.71\% & 4.91\% & 18.21\% & 46.23\% & 60.03\% & 2.48 & 25.33\% \\
\midrule
AnomShield (Ours) & Mistral-7B &0.15\% & 1.53\% & 7.98\% & 26.73\% & 46.48\% & 56.59\% & 6.38 & 40.57\% \\
\bottomrule
\end{tabular}
}
\label{desc-exp}
\end{table*}
\begin{table*}[h]
\centering
\caption{Here we test \textbf{Effect} task on ECVA benchmark using multiple VLMs and the proposed \textbf{AnomShield}. We conduct evaluations on traditional metrics and our \textbf{AnomEval}. It can be seen that AnomShield achieves the \textbf{first place} on AnomEval.}
\resizebox{1.0\linewidth}{!}{
\begin{tabular}{l|c|c|c|c|c|c|c|c|c}
\toprule
\textbf{Method} & BaseLLM &\textbf{BLEU} & \textbf{ROUGE-2} & \textbf{ROUGE-L} & \textbf{BLEURT} & \textbf{MoverScore} & \textbf{UniEval} & \textbf{GPT-Based} & \textbf{AnomEval} \\
\midrule
mPLUG-owl\cite{mplug-owl} & BLoomZ-7B & 0.39\%  & 1.91\% & 12.79\% & 34.32\% & 43.40\% & 49.93\% & 2.95 & 16.65\% \\
\midrule
PandaGPT\cite{su2023pandagpt}& \multirow{5}{*}{LLaMA-1-7B} & 0.22\% & 2.24\% & 11.82\% & 46.42\% & 43.38\% & 47.60\% & 2.35 & 21.10\% \\
Otter\cite{li2023otter}& & 1.07\% & 2.18\% & 14.85\% & 30.52\% & 43.42\% & 46.84\% & 2.20 & 10.72\% \\
ST-LLM\cite{ST-LLM}& & 1.02\% & 2.43\% & 16.66\% & 32.97\% & 43.37\% & 53.89\% & 2.15 & 11.08\% \\
Video-ChatGPT\cite{video-chatgpt}& & 0.28\% & 2.58\% & 11.72\% & 41.03\% & 43.39\% & 44.35\% & 4.28 & 15.70\% \\
A-Guardian\cite{Du_2024_CVPR}& & 0.51\% & 2.60\% & 12.15\% & 42.07\% & 43.28\% & 45.64\% & 4.25 & 15.80\% \\
\midrule
TimeChat\cite{Ren2023TimeChat}& \multirow{5}{*}{LLaMA-2-7B} & 0.76\% & 2.23\% & 13.13\% & 34.34\% & 43.38\% & 40.33\% & 3.25 & 15.19\% \\
LLaMA-VID\cite{llamavid}& & 0.36\% & 3.40\% & 14.42\% & 34.52\% & 43.40\% & 54.01\% & 5.47 & 22.77\% \\
Video-Ilava\cite{videollava}& & 0.33\% & 2.38\% & 11.55\% & 39.98\% & 43.38\% & 51.29\% & 3.71 & 17.17\% \\
MovieLLM\cite{song2024moviellm}& & 0.74\% & 1.44\% & 11.41\% & 30.94\% & 43.38\% & 41.86\% & 1.94 & 21.51\% \\
Chat-Univi\cite{Chat-UniVi}& & 0.55\% & 5.41\% & 18.42\% & 35.57\% & 43.41\% & 55.08\% & 5.74 & 17.65\% \\
\midrule
VILA\cite{vila}&LLaMA-3-8B  & 0.35\% & 3.29\% & 13.42\% & 37.23\% & 43.40\% & 54.29\% & 4.55 & 31.88\% \\
\midrule
Video-LLaMA2\cite{videollama2}&\multirow{2}{*}{Mistral-7B} & 0.12\% & 2.03\% & 8.02\% & 42.85\% & 43.37\% & 50.16\% & 4.42 & 20.21\% \\
VideoChat2-mistral\cite{Videochat2}& & 0.57\% & 2.42\% & 13.60\% & 30.09\% & 43.40\% & 59.20\% & 3.66 & 31.05\% \\
\midrule
AnomShield (Ours)&Mistral-7B & 0.81\% & 3.02\% & 15.67\% & 39.07\% & 43.35\% & 55.78\% & 5.69 & 35.09\% \\
\bottomrule
\end{tabular}
}
\label{outcome-exp}

\end{table*}

\begin{figure*}[!t]
   \centering
    \begin{minipage}{0.333\textwidth}
        \centering
        \subfloat[]{
            \includegraphics[width=0.9\linewidth]{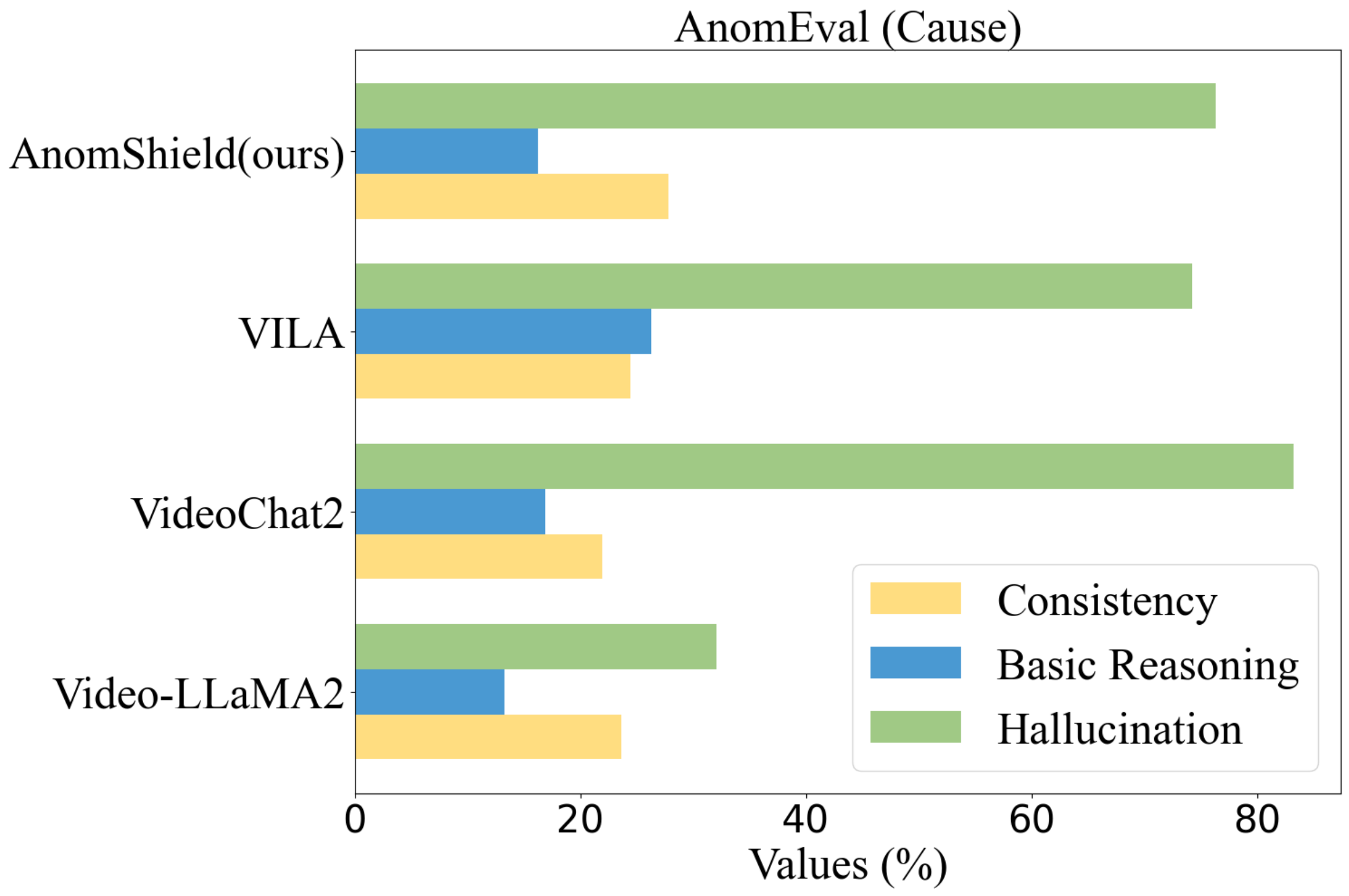}
            \label{fig:exp-Cause}
        }
    \end{minipage}%
    \begin{minipage}{0.333\textwidth}
        \centering
        \subfloat[]{
            \includegraphics[width=0.9\linewidth]{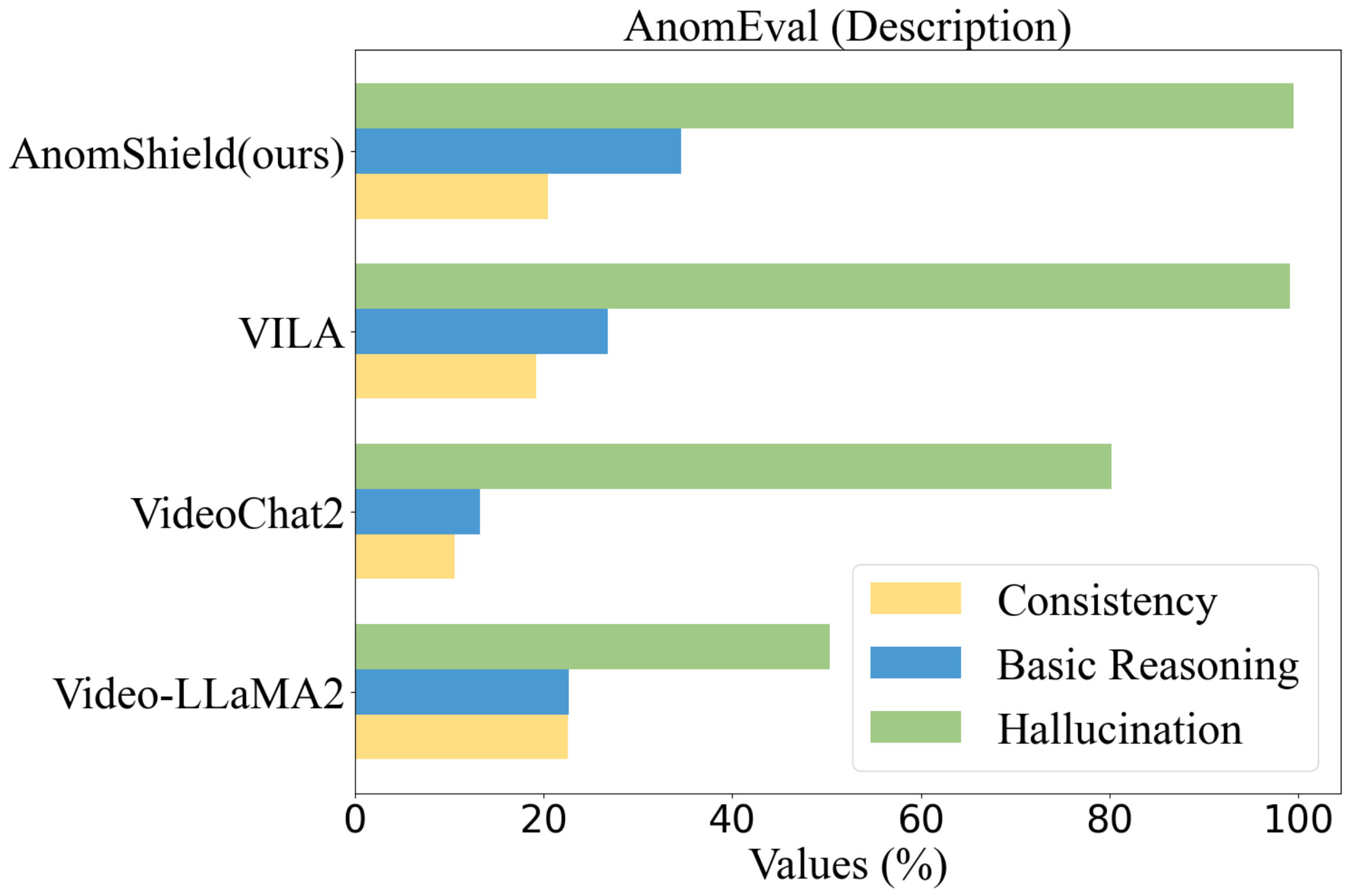}
            \label{fig:exp-Description}
        }
    \end{minipage}%
    \begin{minipage}{0.333\textwidth}
        \centering
        \subfloat[]{
            \includegraphics[width=0.9\linewidth]{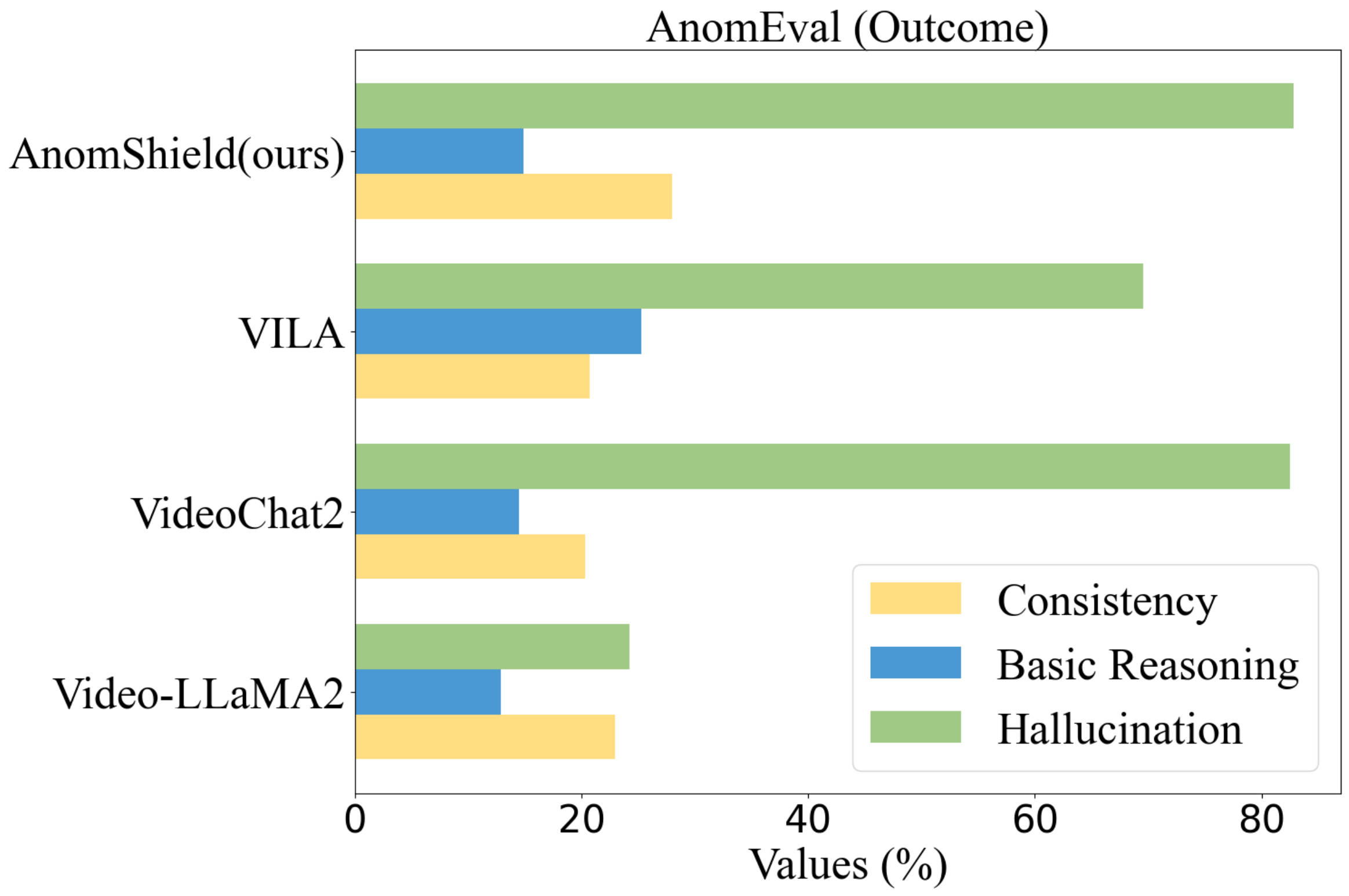}
            \label{fig:exp-Outcome}
        }
    \end{minipage}
    
    \caption{\textbf{Detail comparison of AnomShield.} Figures (a), (b), and (c) demonstrate the performance among the four models (e.g. AnomShield, VideoChat2, VILA, and Videollama2) while evaluating across three different dimensions using the AnomEval. 
 }
    \label{fig:statics}
    \vspace{-5mm}
\end{figure*}
\begin{table}[htbp]

\caption{\textbf{Secondary results on the proposed ECVA benchmark.}  We use the tradition accuracy metrics to evaluate the \textbf{Classification} tasks. We also use IOU to evaluate the \textbf{Timestamp} task, \textbf{N/A} to indicate the model lacks the ability to answer the question.}
\resizebox{0.8\linewidth}{!}{
\begin{tabular}{l|ccc}
\toprule
Methods & Classification & Timestamp \\
\midrule
mPLUG-Owl \cite{mplug-owl}
 & 11.5\%  & 9.0\%\\
LLaMA-VID \cite{llamavid}
& 19.08\%  & 9.17\%\\
PandaGPT \cite{su2023pandagpt}
& 32.6\% &  N/A \\
Otter \cite{li2023otter} 
& 41.3\% & N/A \\
Video-llava \cite{videollava}
& 23.75\% & N/A \\
Video-ChatGPT \cite{video-chatgpt} 
& 21.3\% & 3.20\%\\
TimeChat \cite{Ren2023TimeChat}
& 15.83\% & N/A \\
MovieLLM \cite{song2024moviellm}
& 13.36\% & N/A \\
Video-chat2 \cite{Videochat2}
& 15.21\% & 9.69\% \\
ST-LLM \cite{ST-LLM}
& 18.29\% & N/A \\
Chat-Univi \cite{Chat-UniVi}
& 4.05\% & N/A \\
VILA \cite{vila}
& N/A\% & N/A \\
Video-LLaMa2 \cite{videollama2}
& 46.42\% & N/A \\
\midrule
AnomShield (Ours)
& \textbf{49.59\%} & N/A \\
\bottomrule
\end{tabular}
}
\centering
\vspace{-1mm}
\label{T:Tra-exp}
\end{table}

\begin{table}[htbp!]
\centering
\caption{Here, we conduct the ablation study to validate the efficacy of the ``soft prompt'' and the ``hard promt''.}
\caption{\textbf{Ablation Study}} \label{T:ablation_study}
\centering
\resizebox{0.8\linewidth}{!}{
\begin{tabular}{lccc}
\toprule
\multirow{2}{*}{Model} & \multicolumn{3}{c}{AnomEval ($\%$)} \\
\cmidrule(l){2-4}
& Description & Cause & Effect  \\
\midrule
Ours  & 33.30& 40.57 & 35.09 \\
- Hard Prompt & 31.02& 38.22 & 34.91  \\
- Soft Prompt & 20.23 & 27.57 & 14.12  \\
\bottomrule 
\end{tabular}
}
\vspace{-3mm}
\end{table}

\subsection{Main Results}
\subsubsection{Quantitative evaluation of AnomShield on Causation tasks}
\textbf{Our AnomShield model achieves state-of-the-art performance in both the description and effect tasks.}
We conduct experiments on causation tasks (e.g. Cause, Effect, Description) involved in our dataset. 
For each task, we evaluate the performance of various VLMs and our model using various evaluation metrics.
As shown in Table \ref{cause-exp}, \ref{desc-exp} and Table \ref{outcome-exp}. 
Our model achieves state-of-the-art performance on the description and effect tasks while delivering a secondary result on the cause task.
We attribute this to the efficient architecture of our AnomShield and its advanced training strategies.
Specifically, compared to the secondary method, our AnomShield gains $1.11\%$ and $3.21\%$ improvements in anomaly description and effect understanding, respectively.
In terms of Cause task, VILA achieves the best results, we attribute this reason to the 8B Base LLM architecture adopted by VILA, as well as the fact that their training data is ten times larger than ours.
These outcomes underscore the extraordinary capabilities of AnomShield, showcasing its ability to effectively understand and interpret anomaly events within videos.
All VLMs perform poorly on traditional evaluation metrics, as these metrics solely assess model performance based on the similarity of texts. 
Some other evaluation metrics, such as BLEURT and UniEval which are designed to measure semantic similarity, also struggle to effectively differentiate the performance of various VLM models.
Our model has achieved comparable results on both the AnomEval and GPT-base evaluation metrics.
However, the GPT-based approach tends to favor answers that are more content-rich, which may compromise the assessment of the answers' correctness, and these distinctions are clearly observable through the case study \ref{subsec:case study}.
We also conduct a comparative analysis of AnomShield's performance across multiple dimensions with other models in Figure \ref{fig:statics}. 
\begin{figure*}[hb!]
    \vspace{-2mm}
    \centering
    \includegraphics[width=\linewidth]{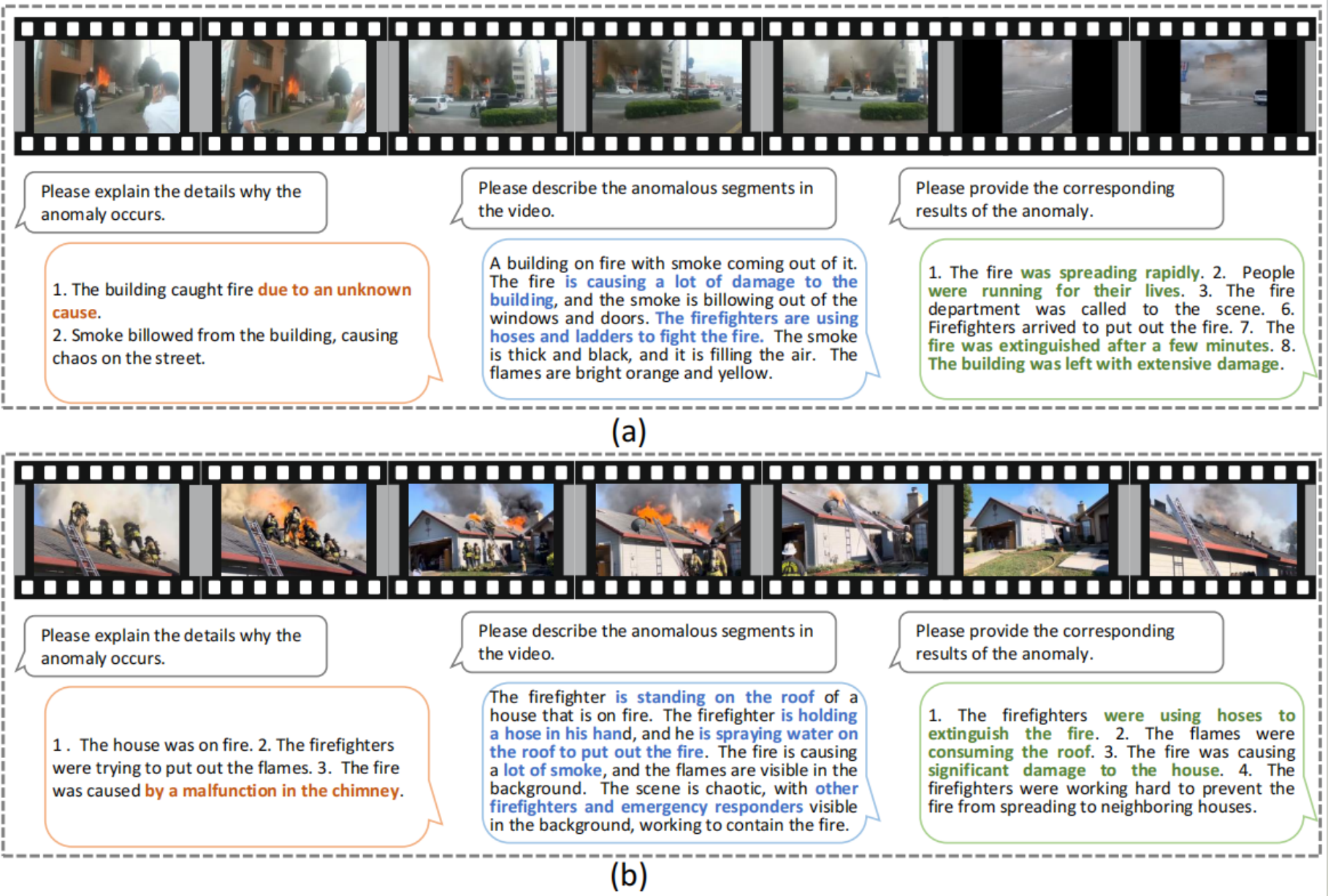}
    \caption{\textbf{Case study}. Here, we demonstrate the performance of our method through two cases along with the model's answers regarding ``What, Why and How'' of the anomaly. It can be seen that our method excels in video anomaly understanding.}
    \label{fig:case_study_model}
\end{figure*}
\subsubsection{Quantitative evaluation of AnomShield on Traditional tasks} 
For the traditional tasks(e.g. Classification, Timestamp), we use a standardized prompt and apply string matching to extract relevant answers from the inference results of VLMs. 
Accuracy metrics are used to evaluate the classification task, while the Intersection Over Union (IOU) metric is applied for the timestamp task.
As shown in Table \ref{T:Tra-exp}, our AnomShield model achieves the highest performance in the classification task. 
We attribute this superiority to the ``hard prompt'' approach, which may cause the model to ignore some details on irrelevant parts of video anomalies. 
\subsubsection{Detail comparison of AnomShield}
To conduct an in-depth analysis of the model's performance, we decompose the AnomEval scores into three dimensions (e.g. Consistency, Basic Reasoning, and Hallucination).
We select several models that perform well in the causation task and compare them with AnomShield.
As shown in Figure \ref{fig:exp-Cause}, Figure \ref{fig:exp-Description} and Figure \ref{fig:exp-Outcome}, we can observe that in all tasks, both AnomShield and VILA demonstrated excellent reasoning capabilities. 
In terms of matching score, AnomShield achieves the best results in both the Cause and Effect tasks, but underperforms in the description task compared to VideoLLama2.
We attribute this advantage to VideoLLama2's integration of the audio modality, allowing it to better describe video segments.
In the Logic Score dimension, VILA performs exceptionally well, which can be attributed to its extensive training data.
\begin{figure*}[h!]
    \centering
    \includegraphics[width=\linewidth]{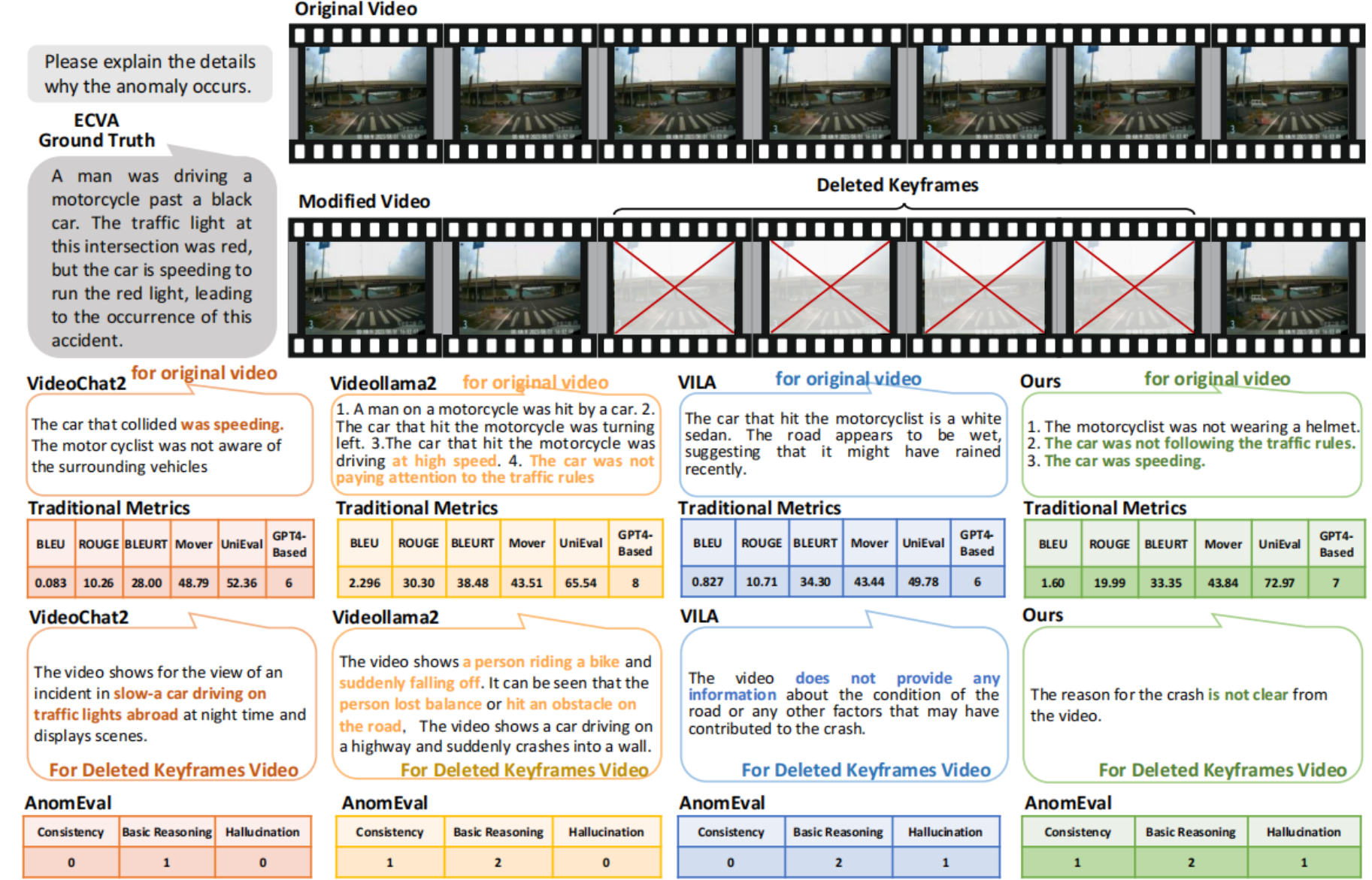}
    \caption{\textbf{Case study}. 
    Here we detail the evaluation process of AnomEval. We compare the responses from four VLMs (e.g. VideoChat2, VideoLLama2, VILA, AnomSheild) and their evaluation results across different metrics. Combining the video content and ground truth, AnomSheild's response is the most consistent with the video content, and our AnomEval is able to provide the accurate and comprehensive results.}
    \label{fig:case_study_eval}
    \vspace{-5mm}
\end{figure*}
\subsection{Ablation Study}
\textbf{Both hard and soft prompts significantly improve the VLM's understanding of the video's causation.}
This section investigates the influence of soft prompts and hard prompts on our method.
Specifically, during the hard prompt ablation experiments, we adjust the sampling strategy from dense sampling to uniform sampling. 
During the soft prompt ablation experiments, we replace the connector part of the model with a two-layer (MLP) network.
As shown in Table \ref{T:ablation_study}, the design of the soft prompt achieves a greater improvement than that of the hard prompt, indicating that the structure of the VLM is more effective in uncovering VLM's reasoning capabilities compared to the hard prompts.

\subsection{Case Study}\label{subsec:case study}
In Figure \ref{fig:case_study_model}, we present several video examples to validate the effectiveness of AnomShield.
As the first case shows, the firefighters are extinguishing a house that is emitting thick smoke. 
Since the video does not display the cause of the fire, the model responds with ``Unknown cause''. 
In the task of anomaly description, the model recognizes a house with ``smoke '' and ``firefighters using ladders and hoses to put out the fire''. 
Regarding the effect task, the model identifies that ``The fire was extinguished'' and at the same time, ``The building was left with extensive damage.''
In the second video, it can be seen that the roof emitted thick smoke, which then suddenly caught fire, so the model responds with ``by a malfunction in the chimney'' for the causation task.
In the anomaly description task, Anomshield recognizes that the firefighter ``is spraying water on the roof to put out the fire''.
From the last few frames of the video, it can be seen that the roof has been charred by the fire, with several areas damaged by the flames.
Thus, in the effect task, the model responds with ``The fire was causing significant damage to the house''.

In Figure \ref{fig:case_study_eval}, we illustrate the performance of VideoChat2, Video-Llama2, VILA and AnomShield, showcasing the different answers they provide for the anomaly causation task.
From this case, we can observe that the main cause of the car accident was the white vehicle violating the traffic signal (e.g. running a red light). 
In addition, speeding by the white vehicle was a secondary factor contributing to the accident. 
When comparing the responses of different models, VideoChat2, VideoLLama2, and AnomShield all correctly identified 'speeding' as a factor. 
Notably, AnomShield and Videollama2 also include the violation of traffic rules, which was not evident in the responses of the other models. 
However, no model accurately pointed out the key fact that the white vehicle ran the red light.
During the evaluation, since VideoLLama2, and AnomShield correctly identified ``speeding'' and  ``traffic rules'' as key factors, they received a perfect matching score of $1$ point. 
In terms of basic reasoning score, the Videollama2, VILA, and AnomShield responses included both the elements of ``car'' and ``motorcyclist''. 
As a result, these three models received a basic reasoning score of $2$ points. 
However, VideoChat2 was slightly inferior in providing critical information, which led to a relatively lower logic score. 
In terms of hallucination, VideoLLama2 and VideoChat2 still provided responses when processing the edited videos, indicating that they may have some hallucination issues, thus receiving a lower reasoning score.

Regarding the evaluation metrics, it can be observed that several assessment criteria measuring the correctness of answers from the perspective of text similarity are inaccurate. 
Meanwhile, it can be observed that the evaluation metrics based on semantic similarity gave similar scores to each answer, indicating that these methods lacks sufficient discriminative power and are not accurate enough.
As the GPT-based evaluation metricd's preference for longer text answers, it gave the highest score to Videollama2's response. 
In contrast, AnomEval can accurately and comprehensively evaluate the capabilities of VLMs, which is more reliable and effective.
\subsection{Result Discussion}
Through experiments, we have discovered and summarized the following conclusions: 
(1) For free-text tasks, most VLMs excel in the description of anomalies but perform poorly on the causation task.
This is because the tasks of description only require the VLM to comprehend the content of the videos, but causation analysis requires the VLM to possess a certain level of reasoning capability to build a logic chain of the cause-effect.
(2) Timestamp localization task is most challenging.
Due to the relatively simplistic temporal and spatial relationships between video frames, VLM performs poorly on fine-grained tasks such as timestamp localization but excels in coarse-grained tasks such as anomaly classification. 
(3) The performance of VLMs is positively correlated to the performance of their underlying large language models.
(4) Training image data is important.
During the training process of the video large language model, relevant high-quality images or videos play an important part in significantly accelerating the training process.
(5) Precise causal analysis often requires additional background knowledge. For example, in the case study of the evaluation process, the VLMs need to understand that a red light means vehicles must stop, and this information must be connected to the traffic accident in the video to construct a causal logic chain. Only in this way can the model accurately determine that the white car caused the accident by running the red light.
\section{Conclusion} \label{sec:Conclusion}
This paper presents ECVA, a novel benchmark for causation understanding of video anomaly. 
To the best of our knowledge, our ECVA is the first benchmark in the field.
Compared with the existing datasets, ECVA is more comprehensive and more challenging with much higher-quality annotations.
We believe the proposed ECVA will encourage the exploration and development of various downstream tasks such as anomaly detection, anomaly prediction, anomaly reasoning, etc. 
we also present a prompt-based approach ``AnomShield'' that can serve as a baseline approach for ECVA. 
Such an approach can capture the key cues of anomalies and build a logic chain of the cause-effect.
Furthermore, we put forward AnomEval, a novel evaluation metric to measure the challenging ECVA in a more comprehensive and robust manner.
Experimental results show that ECVA enables us to develop and evaluate various VLM methods. In the future, we plan to apply ECVA to more scenarios for the challenging video anomaly understanding. 
\vspace{-7pt}
\bibliographystyle{IEEEtran}
\bibliography{main}
\begin{IEEEbiographynophoto}
{Hang Du} received his M.S. degree in computer science from Beijing University of Posts and Telecommunications, Beijing, China, in 2022. He is currently pursuing his Ph.D degree at the National Engineering Research Center for Mobile Network Technologies, Beijing University of Posts and Telecommunications, Beijing. His research interests include multimodal large language model, Semantic Communication, and AI Robustness. He has published papers in top-tier conferences including CVPR and AAAI.
\end{IEEEbiographynophoto}
\vspace{-40pt}
\begin{IEEEbiographynophoto}
{Guoshun Nan} (Member, IEEE) is a tenure-track professor at the National Engineering Research Center for Mobile Network Technologies, Beijing University of Posts and Telecommunications. He has broad interest in natural language processing, computer vision, machine learning, and wireless communications, such as information extraction, model robustness, multimodal retrieval, and next-generation wireless networks. He is a member of the National Engineering Research Center for Mobile Network Technologies. He has published over 20 papers in top-tier conferences and journals, including ACL, CVPR, KDD, EMNLP, SIGIR, IJCAI, CKIM, SIGCOMM, IEEE JSAC, IEEE Communications Magazine, IEEE Network, and Computer Networks. He serves as a reviewer for CVPR, ACL, EMNLP, AAAI, NeurIPS, KDD, IEEE TIP, IEEE TMC and IEEE TIFS. 
\end{IEEEbiographynophoto}
\vspace{-40pt}
\begin{IEEEbiographynophoto}
{Jiawen Qian} is an undergraduate student at the National Engineering Research Center for Mobile Network Technologies, Beijing University of Posts and Telecommunications. He specializes in network security and intrusion detection, aiming to advance his expertise in safeguarding digital systems and detecting cyber threats. 
\end{IEEEbiographynophoto}
\vspace{-40pt}
\begin{IEEEbiographynophoto}
{Wangchenhui Wu}is currently pursuing a bachelor's degree at Beijing University of Posts and Telecommunications. His research interests include NLP and multimodal large language models.
\end{IEEEbiographynophoto}
\vspace{-40pt}
\begin{IEEEbiographynophoto}
{Wendi Deng} received her bachelor's degree in cyberspace security from Beijing University of Posts and Telecommunications, Beijing, China, in 2024.  She is currently pursuing her M.S. degree at the National Engineering Research Center for Mobile Network Technologies, Beijing University of Posts and Telecommunications, Beijing. Her research interests include large language model and multimodal large language model.
\end{IEEEbiographynophoto}
\vspace{-40pt}
\begin{IEEEbiographynophoto}
{Hanqing Mu} received his bachelor's degree in cyberspace security from Beijing University of Technology, Beijing, China, in 2023.  He is currently pursuing his M.S. degree at the National Engineering Research Center for Mobile Network Technologies, Beijing University of Posts and Telecommunications, Beijing. His research interests include large language model and privacy preservation.
\end{IEEEbiographynophoto}
\vspace{-40pt}
\begin{IEEEbiographynophoto}
{Zhenyan Chen} received her bachelor's degree in cyberspace security from Beijing University of Posts and Telecommunications, Beijing, China, in 2024.  She is currently pursuing his M.S. degree at the National Engineering Research Center for Mobile Network Technologies, Beijing University of Posts and Telecommunications, Beijing. Her research interests include large language model and multimodal large language model.
\end{IEEEbiographynophoto}
\vspace{-40pt}
\begin{IEEEbiographynophoto}
{Pengxuan Mao} received his Ph.D. degree from Beijing Jiaotong University, He is currently a senior researcher at Terminus group., Ltd. Beijing .  His main areas of interest are computer vision and IoT.
\end{IEEEbiographynophoto}
\vspace{-40pt}
\begin{IEEEbiographynophoto}
{Xiaofeng Tao} (Senior Member, IEEE) received the bachelor’s degree in electrical engineering from Xi’an Jiaotong University, Xi’an, China, in 1993, and the master’s and Ph.D. degrees in telecommunication engineering from the Beijing University of Posts and Telecommunications (BUPT), Beijing, China, in 1999 and 2002, respectively. He is currently a Professor at the National Engineering Research Center for Mobile Network Technologies, Beijing University of Posts and Telecommunications, a fellow of the Institution of Engineering and Technology, and the Chair of the IEEE ComSoc Beijing Chapter. He has authored or coauthored over 200 articles and three books in wireless communication areas. He focuses on B5G/6G research.
\end{IEEEbiographynophoto}
\vspace{-40pt}
\begin{IEEEbiographynophoto}
{Jun Liu} (IEEE Senior Member) is a Professor and Chair in Digital Health at School of Computing and Communications in Lancaster University. He was with Singapore University of Technology and Design from 2019 to 2024. He is an Associate Editor of IEEE Transactions on Image Processing, IEEE Transactions on Industrial Informatics, IEEE Transactions on Biometrics, Behavior and Identity Science, ACM Computing Surveys, and Pattern Recognition. He has served as an Area Chair of CVPR, ECCV, ICML, NeurIPS, ICLR and MM. His research interests include computer vision, machine learning and digital health.
\end{IEEEbiographynophoto}
\end{document}